\documentclass[11pt]{article}

\usepackage[final]{acl}

\usepackage{times}

\usepackage[T1]{fontenc}

\usepackage[utf8]{inputenc}

\usepackage{microtype}

\usepackage{inconsolata}

\usepackage{graphicx}
\usepackage{amsmath}
\usepackage{enumitem}
\usepackage{tikz}
\usetikzlibrary{arrows.meta,positioning,fit,backgrounds,calc}
\usepackage{booktabs}
\usepackage{algorithm}
\usepackage{algpseudocode}
\usepackage{tabularx}

\usepackage{xspace}

\definecolor{genBlue}{HTML}{0072B2}
\definecolor{frameOrange}{HTML}{E69F00}
\definecolor{genericGreen}{HTML}{009E73}

\newcommand{\generalised}[1]{\textcolor{genBlue}{#1}}
\newcommand{\framing}[1]{\textcolor{frameOrange}{#1}}
\newcommand{\genericity}[1]{\textcolor{genericGreen}{#1}}

\newcommand{\labsep}{\texttt{/}\allowbreak}

\newcommand{\geng}{%
  \texttt{\generalised{gen}}\labsep
  \texttt{\framing{bare}}\labsep
  \texttt{\genericity{generic}}\xspace%
}

\newcommand{\genglong}{%
  \texttt{\generalised{generalised}}\labsep
  \texttt{\framing{bare}}\labsep
  \texttt{\genericity{generic}}\xspace%
}

\newcommand{\genfg}{%
  \texttt{\generalised{gen}}\labsep
  \texttt{\framing{framed}}\labsep
  \texttt{\genericity{generic}}\xspace%
}

\newcommand{\genfglong}{%
  \texttt{\generalised{generalised}}\labsep
  \texttt{\framing{framed}}\labsep
  \texttt{\genericity{generic}}\xspace%
}

\newcommand{\genfng}{%
  \texttt{\generalised{gen}}\labsep
  \texttt{\framing{framed}}\labsep
  \texttt{\genericity{non-generic}}\xspace%
}

\newcommand{\genfnglong}{%
  \texttt{\generalised{generalised}}\labsep
  \texttt{\framing{framed}}\labsep
  \texttt{\genericity{non-generic}}\xspace%
}

\newcommand{\geno}{%
  \texttt{\generalised{gen}}\labsep
  \texttt{\framing{bare}}\labsep
  \texttt{\genericity{non-generic}}\xspace%
}

\newcommand{\genolong}{%
  \texttt{\generalised{generalised}}\labsep
  \texttt{\framing{bare}}\labsep
  \texttt{\genericity{non-generic}}\xspace%
}

\newcommand{\ngen}{%
  \texttt{\generalised{non-gen}}\xspace%
}

\newcommand{\ngenlong}{%
  \texttt{\generalised{non-generalised}}\xspace%
}

\newcommand{\genast}{%
  \texttt{\generalised{gen}}\labsep\texttt{*}\xspace%
}

\newcommand{\genastlong}{%
  \texttt{\generalised{generalised}}\labsep\texttt{*}\xspace%
}

\newcommand{\genfast}{%
  \texttt{\generalised{gen}}\labsep
  \texttt{\framing{framed}}\labsep
  \texttt{*}\xspace %
}

\newcommand{\framed}{%
  \texttt{*}\labsep
  \texttt{\framing{framed}}\labsep
  \texttt{*}\xspace %
}

\newcommand{\generic}{%
  \texttt{*}\labsep
  \texttt{\genericity{generic}}\xspace %
}

\newcommand{\nongeneric}{%
  \texttt{*}\labsep
  \texttt{\genericity{non-generic}}\xspace %
}

\newcommand{\fnsep}{\texttt{\_}\allowbreak}

\newcommand{\aboutoutcome}{\texttt{about}\fnsep\texttt{outcome}\xspace}
\newcommand{\isgen}{\texttt{is}\fnsep\texttt{generalised}\xspace}
\newcommand{\gettargetsent}{\texttt{get}\fnsep\texttt{target}\fnsep\texttt{sent}\xspace}
\newcommand{\refstudy}{\texttt{ref}\fnsep\texttt{study}\xspace}
\newcommand{\refstudypop}{\texttt{ref}\fnsep\texttt{study}\fnsep\texttt{population}\xspace}
\newcommand{\ispresenttense}{\texttt{is}\fnsep\texttt{present}\fnsep\texttt{tense}\xspace}
\newcommand{\isreporting}{\texttt{is}\fnsep\texttt{reporting}\fnsep\texttt{clause}\xspace}
\newcommand{\containsembedded}{\texttt{sentence}\fnsep\texttt{embedded}\xspace}
\newcommand{\containsquant}{\texttt{quantified}\xspace}
\newcommand{\containsvague}{\texttt{is}\fnsep\texttt{vague}\xspace}
\newcommand{\containshedging}{\texttt{contains}\fnsep\texttt{hedging}\xspace}

\newcommand{\removableframe}{\texttt{removable}\fnsep\texttt{framing}\xspace}

\newcommand{\datasetname}{\textsc{NLPGens}\xspace}
\newcommand{\taxonomyname}{\textsc{NLPGenX}\xspace}
\newcommand{\frameworkname}{\textsc{NLPGenA}\xspace}

\definecolor{MyColour}{RGB}{87, 66, 114}
\definecolor{GenericFramedColour}{RGB}{217, 136, 89}

\usepackage{tcolorbox}
\tcbuselibrary{breakable}
\newtcolorbox[auto counter,number within=section]{prompt}[1][]{
    colbacktitle=black!60,
    coltitle=white,
    fontupper=\footnotesize,
    boxsep=5pt,
    left=0pt,
    right=0pt,
    top=0pt,
    bottom=0pt,
    boxrule=1pt,
    title={#1},
    breakable,
    #1,
}

\title{How broad is that claim? Mapping Generalisation in NLP Research}

\author{Chenxin Diao \And 
Nataliya Stepanova \\ 
School of Informatics, University of Edinburgh \\
  \texttt{\{chenxin.diao,nstepano,emily.allaway\}@ed.ac.uk}\\
  \And Emily Allaway\\
  }

\begin{document}
\maketitle
\begin{abstract}
Generalisations are common in scientific communication, even though they are semantically ambiguous.
An automated method is needed to identify and categorise claims according to their level of generalisation, in order 
help detect an over-reliance on generalisations and possible misrepresentations of scientific findings.
We introduce a comprehensive taxonomy of generalisations in the scientific domain, \textbf{\taxonomyname}, which labels claims according to their level of generality and framing within the text. We operationalise this taxonomy with an LLM-powered framework, \textbf{\frameworkname}, that automatically classifies sentences from scientific articles into 
5 different generalisation classes.
We validate our framework with human annotators and use the framework to construct
a large-scale dataset of NLP papers annotated according to generality, with auxiliary labels for hedging and vague descriptors (\textbf{\datasetname}).
We use \datasetname to analyse the use of generalisations in NLP papers across multiple venues and subdomains, and to examine associations with citation counts, hedging, and vague descriptors.

The dataset and code are available at \url{https://github.com/cx-diao/nlpgen}.
\end{abstract}

\section{Introduction}

Generalisations allow
researchers to extrapolate from concrete findings to broader implications~\citep{bunge1960place,peters2022generalization}.
As a result, they are abundant in scientific communication, including in research articles~\citep{dejesus2019generic}.
However, by obscuring variability, these generalisations may also
exaggerate or overgeneralise findings beyond what is actually supported by evidence, leading to potential misinterpretations.

\begin{figure}[!t]
\centering
\includegraphics[width=\columnwidth]{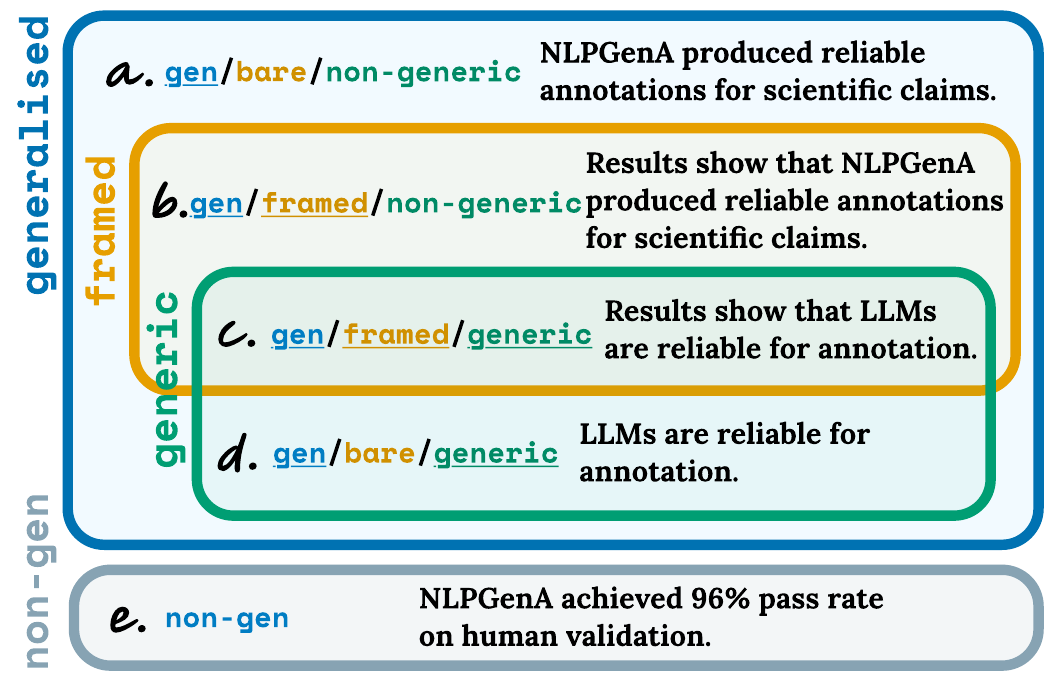}
\caption{Our taxonomy \taxonomyname for classifying outcome-reporting sentences in the NLP domain.}
\label{fig:substantiator_gen_categories}
\end{figure}

Generalisations can take many forms in language (see Fig~\ref{fig:substantiator_gen_categories}), with varying semantic interpretations~\citep{haigh2020does, bowker2023problem}. 
Notably, generics express generalisations without qualifying how broadly they apply.
In both humans \citep{leslie2011all} and large language models (LLMs)~\citep{peters2025generics,allaway2024exceptions}, this has been shown to lead to overgeneralisation of a conclusion.
While generalisations, particularly generics, have been extensively studied~\citep[e.g.,][]{Krifka1995,Leslie2008, carlson2010generic, neufeld2025giving}, most studies focus on relatively simple sentences. The resulting theoretical definitions are then either too abstract or syntactically limited (e.g., to only bare plural subjects) to translate directly to the scientific domain.
Therefore, we propose a new taxonomy that captures differing forms of generalisations in science. 

Our taxonomy (\textbf{\taxonomyname}) unifies definitions from across multiple recent studies on generalisations in science \citep{dejesus2019generic, peters2024hasty_medical, peters2024hasty_philosophy}.
Focusing on 
sentences about results from
the NLP domain, we define \textbf{five categories} that capture different degrees of generalisation.
In scientific articles, very general claims are often embedded within study-specific framing (see Fig.~\ref{fig:substantiator_gen_categories}c). 
We draw on \citet{dejesus2019generic} to capture this in our taxonomy.

Automatically identifying and classifying generalisations is challenging due to the complexity of both the definitions and of scientific language.
While prior studies have manually identified generalisations in multiple domains (e.g., medical research, experimental philosophy;~\citealp{peters2024hasty_medical,peters2024hasty_philosophy}),
these relied on extensive manual annotation, limiting their scale.
Additionally, systems to automatically identify generalisations have limited coverage, focusing only on a subset of generalisations that are identifiable through heuristic short-cuts --- e.g., kind-referring plural noun phrases in the subject position \citep{ACE, ACE_2, ACE_2005, ARRAU, mgen}.
To facilitate large-scale automatic analysis, we propose a novel framework that uses LLMs to operationalise our taxonomy in the NLP domain.

We considerably extend on previous automation efforts by providing an LLM-driven framework (\textbf{\frameworkname}) to categorise sentences following our taxonomy.
Our framework uses nine consolidated annotation tasks: seven filter outcome sentences and derive taxonomy labels, while two auxiliary tasks identify hedging and vague descriptors.
We collect human annotations to validate our framework, demonstrating substantial agreement between the LLM annotations and the human experts.

We apply our framework to the 4.5K documents in the NLP subset of the NLPeer dataset~\citep{dycke-etal-2023-nlpeer}, a corpus of research articles in the NLP domain from 2017--2024.
From this we create a new corpus, \textbf{\datasetname}, of 180K sentences filtered and then labelled according to our taxonomy, hedging, and vague descriptors.
We then use \datasetname to study how generalisations are used in NLP literature. We conduct analyses
on how the distribution of different generalisation types varies across venues and NLP sub-domains.
We also conduct exploratory analyses into paper citation counts.
We find that both hedging and vague descriptors are associated with generalisations in NLP papers.

Our contributions are as follows:
\begin{enumerate}[itemsep=0em]
    \item A taxonomy of generalisations in NLP papers, \taxonomyname.
    \item A robust automated annotation framework, \frameworkname, consisting of classifiers to label the sentences according to our taxonomy.
    \item A large-scale dataset, \datasetname, of sentences from NLP papers annotated for levels of genericity, hedging, and vague descriptors, obtained via our framework.
    \item Analyses on generalisations in NLP papers, including their associations with other linguistic features (hedging and vague descriptors) and paper impact (measured through citation counts).
\end{enumerate}

\section{Prior Work}

Generalisations are common across languages \citep{behrens2005genericity, carlson1995thegenericbook}, and are often classified as habitual or causal~\citep{tessler2019language}.
Generic statements are a special subtype of generalisations that omit explicit quantification.
They have garnered significant interest as they can be judged true based on weak evidence but have strong implications \citep{cimpian2010generic}.
Usage of generalisations and generics has implications for science, as the language used for communicating findings impacts the conclusions people draw \citep{peters2025generics}.

\paragraph{Labelling generalisations in language.}
Early annotation schemas for generalisations 
were restricted to labelling the genericity of noun phrases~\citep{ACE, ACE_2, ACE_2005}. Later works also labelled the genericity of a clause, using event descriptions and co-reference taxonomies~\citep{ECB+, RED, friedrich2015annotating, SitEnt} and classified events as \textit{episodic} or \textit{habitual} \citep{krifka1995genericity, mathew2009supervised, smith2003modes}. 
More recently, \citet{govindarajan2019decomposing} presented a semantic framework for labelling predicates and their arguments.
In contrast to these works, we categorise and annotate generalisations at the \textit{sentence} level.

Initial 
automated approaches explored the use of syntactic and grammatical features for annotating generalisations~\citep{friedrich2015automatic,friedrich2015discourse}.
More recent works have fine-tuned language models~\citep{devlin2019bert} for automatic annotation~\citep{bhakthavatsalam2020genericskb,mgen}. 
However, these approaches require labelled data for training, which is difficult to obtain. So, we use zero-shot prompting with an instruction-tuned LLM for our annotation framework.

\paragraph{Analysing scientific articles.}
The claims in scientific articles have been analysed for a number of aspects including discourse role~\citep{fisas-etal-2015-discoursive}, citation purpose~\citep{fisas-etal-2016-multi}, and argumentation relations~\citep{lauscher-etal-2018-investigating}. 
Prior works have also studied the relationship between scientific claims and news headlines about them~\citep{wuehrl-et-al_2024understanding}. These aspects relate primarily to the pragmatics of a claim and its role or relationships. In contrast, we focus on the semantics of claims by labelling their generality.

Manual annotations have been conducted to label
generalisations in scientific articles from psychology~\citep{dejesus2019generic}, medicine~\citep{peters2024hasty_medical}, and experimental philosophy~\citep{peters2024hasty_philosophy}.
While these labelling schema had overlapping categories, such as hedging and framing, their definitions often included domain-specific elements or were designed to cover only a limited set of claims (e.g., only titles and abstracts).
We define a novel taxonomy for generalisations that draws on these works while also providing concrete, actionable definitions.

\section{\taxonomyname: A Taxonomy of Scientific Outcome Generalisations}
\label{sec: theory}

\begin{figure*}
    \centering
    \includegraphics[width=\linewidth]{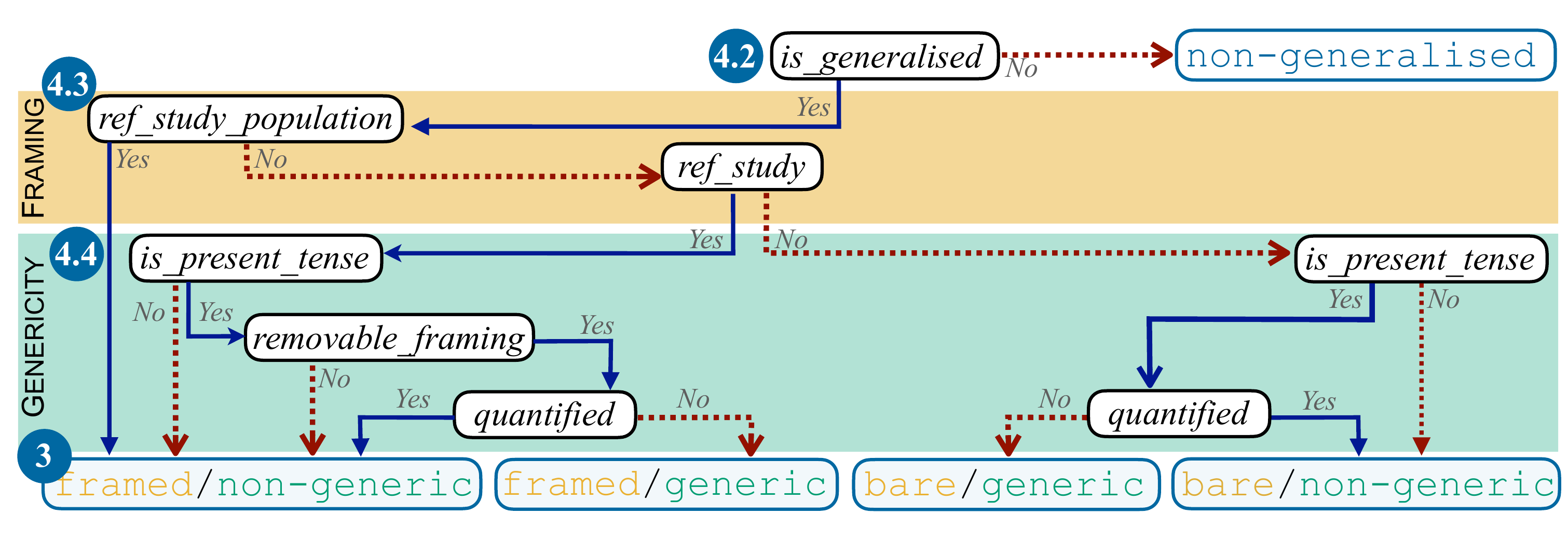}
    \caption{Our framework for annotating generalisations in NLP papers (\S\ref{sec:framework}) using our proposed taxonomy (\S\ref{sec: theory}).}
    \label{fig:generalised-decision-tree}
\end{figure*}

Generalisations are claims whose interpretation supports inference beyond the particular observations explicitly reported in the sentence.
Here we focus specifically on generalisations about the findings (outcomes) of a study.
We categorise these
generalisations along 
two dimensions (\S\ref{subsec:flags}):
framing and genericity. 
\textbf{Genericity} concerns how broadly a claim is interpreted as applying. \textbf{Framing} concerns how the generalisation is presented in relation to the study. 
These dimensions draw on prior work in psychology, medicine, and experimental philosophy~\citep{dejesus2019generic,peters2024hasty_medical,peters2024hasty_philosophy}, which we unify into our taxonomy.
Their recurrence across fields suggests that \taxonomyname may extend beyond NLP, but this requires domain-specific validation (see \S\ref{sec:limitations}).

\subsection{Dimensions of Generalisation}
\label{subsec:gendims}

\paragraph{Framing.}
Inspired by \citep{dejesus2019generic}, we take into account the framing of generalisations in our taxonomy.
Framing is central to generalisation in scientific communication. 
Authors often 
embed a generalisation inside a reporting frame, such as ``our results suggest'' or ``this study demonstrates''. This frame
then links
the generalisation to the particular study (Fig,~\ref{fig:substantiator_gen_categories}c and d).
Framing may also take the form of references to specific tables or figures in the article. 

\paragraph{Genericity.}
Generalisations can be either generic or non-generic. 
Generic generalisations lack syntactic features (e.g., quantification) that signal how broadly they apply (e.g., Fig.~\ref{fig:substantiator_gen_categories}b). As a result, generics are often interpreted overgenerally as holding broadly (or even universally)~\citep{leslie2011all}.
In contrast, non-generic generalisation include language that qualifies their scope, such as quantification (e.g., ``\frameworkname is mostly reliable'') or references to specific instances.

\subsection{Taxonomic Categories}
We define five categories derived from the different types of generality and framing.
The first level of our taxonomy differentiates between \ngenlong and \genastlong claims, distinguished by how broadly they are implied to apply. 
Non-generalisations mention only specific results in the claim (Fig.~\ref{fig:substantiator_gen_categories}e); 
whereas generalisations 
are implied to apply more broadly (e.g., Fig.~\ref{fig:substantiator_gen_categories}b). 
The second level (\framed) marks whether the scope of a generalisation is constrained through syntactic framing.
Finally, the third level distinguishes \generic versus \nongeneric generalisations.

\paragraph{\genglong.} Bare generic generalisations
are 
claims that express generalisations without any overt study frame nor explicit cues as to how broadly they apply (Fig.~\ref{fig:substantiator_gen_categories}d).

\paragraph{\genfglong.} Framed generic generalisations consist of a generic claim embedded within explicit study framing. These claims present a broadly expressed finding within the context of the study (Fig.~\ref{fig:substantiator_gen_categories}c). 

\paragraph{\genfnglong.} 
Framed non-generic generalisations 
present a finding as applying beyond specific instances but within the context of the study (e.g., ``Results'' in Fig.~\ref{fig:substantiator_gen_categories}b). These claims also include explicit markers of how broadly the claim applies
(e.g., Fig.~\ref{fig:substantiator_gen_categories}b applies to scientific claims). 

\paragraph{\genolong.} Bare non-generic generalisations 
constrain the scope of the claim through means other than framing. That is, a constrained generalisation is presented without explicitly being presented as a finding of the current study (Fig.~\ref{fig:substantiator_gen_categories}a). 

\paragraph{\ngenlong.} Non-generalised claims describe specific results or instances without drawing general conclusions about them 
(Fig.~\ref{fig:substantiator_gen_categories}e).

\section{\frameworkname: A Framework for Annotating Generalisations in NLP}
\label{sec:framework}
We propose a framework to annotate generalisations in NLP papers according to our proposed taxonomy (see \S\ref{sec: theory}).
The framework uses \textbf{nine consolidated annotation tasks}: seven filter outcome sentences and derive taxonomy labels, while two auxiliary tasks identify hedging and vague descriptors respectively.
Since our taxonomy is specifically designed for
claims about outcomes, 
the first step of our framework labels sentences
for this 
(\S\ref{subsec:isoutcome}). 
Then, we identify generalisations (\S\ref{subsec:isgen}) and, for all generalisations, determine whether they are framed in reference to the study (\S\ref{subsec:isframe}) and whether they contain generic language (\S\ref{subsec:isgeneric}).
Independently of the taxonomy, we annotate hedging and vague descriptors (\S\ref{subsec:flags}).
Figure~\ref{fig:generalised-decision-tree} shows our framework.
See Appendix~\ref{app:implementation} for more details.

Appendix~\ref{app:rule-baseline} evaluates a deterministic rule-based counterpart to explore which annotation decisions require LLM-based semantic judgement.
We find that although some of the annotation decisions can be made with deterministic rules, there are significant gaps in performance for annotation tasks that require semantic judgement.
For consistency, we use LLM-based classification for all experiments, and we discuss the potential for rule-based classification in Section~\ref{sec:limitations}.

\subsection{Is it about outcomes?}
\label{subsec:isoutcome}
The first step of our framework determines whether a sentence is about the outcomes of the study. Outcomes include not only experimental results and findings but also other scientific artefacts such as datasets and leaderboards.
Following 
the definitions 
from 
\citet{fisas-etal-2015-discoursive}, we label a sentence as being about outcomes if it is primarily about the author's own work and:
\begin{itemize}[noitemsep,leftmargin=*]
    \item Introduces or presents some scientific artefact (e.g., ``We present...'').
    \item Describes, interprets, or analyses experimental results (e.g., ''The results show...'').
    \item Describes how the research will contribute to knowledge in the field. 
\end{itemize}

\subsection{Is it generalised?}
\label{subsec:isgen}
We label a claim as \genast if it meets at least one of three conditions, combined in \textbf{one classifier (\isgen)}: 
\begin{itemize}[noitemsep, leftmargin=*]
    \item Underspecified quantification (e.g., ``most'', ``often''). We exclude quantifiers with specific thresholds (e..g, ``50\% of'', ``all'') because these refer to specific instances and therefore do not make generalisations.
    \item Expressions of causality (e.g., ``results in'').
    \item Only a minor change in meaning, in context, after the insertion ``usually'' or ``generally''.  
\end{itemize}
The first two conditions capture quantified, habitual, or causal generalisations in line with the distinctions made in prior work~\citep{tessler2019language}. The last condition captures potentially generic generalisations following a test proposed by \citet{krifka1995genericity}.

\subsection{Is it framed?}
\label{subsec:isframe}
If a claim is generalised, we use \textbf{two classifiers} to determine whether it is framed. A \framed claim should refer to the study in some way (\S\ref{subsec:gendims}). This can be through a reference to the study population (\textbf{\refstudypop}), the broader domain most closely related to the the specific instances that were studied (e.g., ``annotators'', ``tested datasets''). Or it can be through more general references to the study (\textbf{\refstudy}), such as mentioning the authors (e.g., ``we find''), the study results (e.g., ``The results show''), or tables and figures (e.g., ``see Table N'').
Claims without a reference to the study are unframed.

\subsection{Is it generic?}
\label{subsec:isgeneric}
Following \citet{krifka1995genericity,dejesus2019generic}, we define a generic claim (using \textbf{three classifiers}) as meeting all of the following conditions
\begin{itemize}[noitemsep,leftmargin=*]
    \item Is in the present tense (\textbf{\ispresenttense}): as noted in \citet{dejesus2019generic}, past tense restricts the scope of a claim to the specific instances from the study.
    \item Does not have quantification (\textbf{\containsquant}): this is a hallmark of generic sentences~\citep{krifka1995genericity} because quantifiers signal how broadly a claim applies.
    \item Does not refer to the study population (\S\ref{subsec:isframe}): framing may reference the study itself but not the sample or population because this constrains the scope of the claim.
\end{itemize}
Unframed claims that meet these conditions are \geng. For framed claims, the 
framing is removed (\textbf{\removableframe}) before determining genericity. 
This is done by determining whether the
study framing can be separated from a complete embedded sentence\footnote{We use three separate prompts for this but aggregate them for simplicity. See Appendix~\ref{app:removable-framing-validation} for full details}. For example, ``our results show that \textit{attention is useful}'' can be (and in fact is) a \genfg; in contrast, ``our results show \textit{the effectiveness of our method}'' can not be generic because removing the framing leaves only a nominal phrase.
If the embedded sentence meets the conditions of genericity, then it is a \genfg.

\subsection{Hedging and Vague Descriptors}
\label{subsec:flags}
We use two classifiers to annotate auxiliary linguistic attributes independently of the taxonomy: hedging (\textbf{\containshedging}) and vague descriptors (\textbf{\containsvague}).
Neither attribute contributes to the final taxonomy label.

Hedging is the use of linguistic expressions (e.g., ``likely'') to show that the author is not fully committed to a claim~\citep{lakoff1973hedges}.
It is common in academic writing, where authors use it to temper certainty and allow for alternatives and criticism~\citep{HYLAND1994239}.
We allow both generic and non-generic claims to be labelled as hedged, following \citet{dejesus2019generic}.

We define a vague descriptor as an under-specified adjective or adverb whose degree is left unclear, such as ``significant'', ``good'', or ``efficient'' without an explicit threshold.

\begin{table}[!t]
\centering
\small
\begin{tabular}{l c c c c}
\toprule
 venue &
  \# papers &
  \# sents &
  \begin{tabular}[c]{@{}c@{}}\# section\\ sents\end{tabular} &
  \begin{tabular}[c]{@{}c@{}}\# outcome\\ sents\end{tabular}\\
  \midrule
\textsc{acl} '17      & 133   & 26K  & 8.9K   & 4.1K   \\
\textsc{coling} '20   & 89    & 16K  & 6.3K   & 2.9K   \\
\textsc{arr} '22      & 473   & 97K  & 42,45  & 19K  \\
\textsc{emnlp} '23    & 2.0K & 413K & 133K & 68K  \\
\textsc{emnlp} '24    & 1.7K & 440K & 205K & 87K  \\
\midrule
\textbf{Total}        & \textbf{4.5K} & \textbf{991K} & \textbf{396K} & \textbf{180K} \\
\bottomrule
\end{tabular}
\caption{Statistics of \datasetname.
``Sents'' is sentences. Section sents are sentences from only relevant paper sections (abstract, introduction, results, discussion, conclusion; \S\ref{subsec:imps}). Outcome sents are sentences labelled as being about the study outcomes (\S\ref{subsec:isoutcome}). 
}
\label{tab:dataset_stats}
\end{table}

\section{\datasetname: A Large-Scale Dataset of Generalisations in NLP Research}
\label{sec:dataset}

\subsection{Implementation}
\label{subsec:imps}
\paragraph{Source data.}
We source NLP papers from the NLPeer corpus~\citep{dycke-etal-2023-nlpeer}. We use all NLP subsets of both v1 and v2 of the corpus (see Table~\ref{tab:dataset_stats}). We use keywords to identify and extract all sentences from five standard paper sections: Abstract, Introduction, Results, Discussion, and Conclusion. See Appendix~\ref{app:implementation} for details.

\paragraph{LLM setup.}
We construct a separate prompt for each of the tasks following the questions described in \S\ref{sec:framework} (see Fig.~\ref{fig:generalised-decision-tree})  and query gpt-oss-120b \citep{agarwal2025gpt} with temperature $=0$ three times
\footnote{Since we used \href{https://vllm.ai/}{vLLM} \citep{kwon2023efficient} to host the LLM, even with temperature $=0$, the output is not deterministic.}.
We take the majority vote for each classifier and aggregate following a decision tree to acquire the final LLM label.
For determining whether a sentence is about outcome, generalised, and whether it refers to a study population, we provide the previous two sentences as context in the prompt.
See Appendix~\ref{app:prompt_templates} for full prompts and details.

\subsection{Dataset Statistics}
Our dataset contains $180{,}426$ fully labelled \aboutoutcome sentences distributed across the five taxonomy classes. On average, $51\%$ of
sentences in a paper were labelled as about outcomes (see Appendix Table~\ref{tab:about_outcome_per_section} for specifics).
Each sentence also carries the auxiliary \containshedging and \containsvague annotations.
As shown in Table~\ref{tab:label-dist}, our dataset is dominated by \genfng ($54.6\%$) and \ngen ($31.6\%$), with the three remaining classes together accounting for $13.8\%$ of the labelled sentences. We present detailed analysis in \S\ref{sec:analysis}. 

\begin{table}[t]
    \centering
    \small
    \setlength{\tabcolsep}{4pt}
    \renewcommand{\arraystretch}{1.15}
    \begin{tabular}{@{}>{\raggedright\arraybackslash}p{0.33\columnwidth}>{\raggedright\arraybackslash}p{0.66\columnwidth}@{}}
        \toprule
        \textbf{Final label} & \textbf{Example}                                                                                          \\
        \hline

        \ngen
                             & CrossAligner exceeds the F-Score of the Previous SOTA by 2.7 points (82.5 versus 79.8).                   \\
        \hline

        \genfng
                             & We believe that our entropy-based perspective will help provide a strong starting point for this pursuit. \\
        \hline
        \genfg
                             & We believe the improvement is from the accurate recognition of these sentences.                                                                                                          \\
        \hline
        \geno
                             &  Most researchers do not take advantage of recent metrics that correlate better with human judgments.                                                                                                         \\
        \hline
        \geng
                             & HiTab also presents cross-domain and complicated calculation challenges.                                  \\
        \bottomrule
    \end{tabular}
    \caption{Examples from our NLPGens corpus (\S\ref{sec:dataset}). See Appendix Table~\ref{tab:path-error-analysis} for more examples.}
\end{table}

\begin{table*}[t]
    \centering
    \small
    \begin{tabular}{l|rr|rrrr}
        \toprule
        & \multicolumn{2}{c|}{Full Corpus} & \multicolumn{4}{c}{End-to-End Annotated Subset}\\
        \textbf{Final label} & $\mathbf{n}$ & \textbf{\% label dist} & $\mathbf{R(L)}$  & $\mathbf{n_G}$ & \textbf{correct} & \textbf{acc} \\
        \hline
        \ngen                & 57050            & 31.6  & 1.00 & 29           & 28               & 0.966  \\
        \genfng              & 98449            & 54.6  & 0.95 & 60           & 58               & 0.967 \\
        \geng                & 15852            & 8.8   & 0.94 & 3            & 3                & 1.000 \\
        \geno                & 6956             & 3.9   & 0.92 & 5            & 4                & 0.800  \\
        \genfg               & 2119             & 1.2   & 0.75 & 3            & 3                & 1.000  \\
        \hline
        \textbf{Total}       & \textbf{180426}  & 100.0 & -    & 100          & 96               & 0.960  \\
        \bottomrule
    \end{tabular}
    \caption{Label statistics and annotator validation for our corpus. $n$ is the number (\% label dist is the percentage) of \aboutoutcome sentences in the corpus assigned each label by our framework.
    $n_G$ is the number of sentences for each final label where all 9 classifier labels were validated by humans, \textit{correct} is the number where LLM-label was validated as correct and \textit{acc} is the corresponding LLM accuracy.
    $R(L)$ is data-weighted per-label reliability (Eq.~\ref{eq:label-trust}).
    }
    \label{tab:label-dist}
\end{table*}

\subsection{Human Validation Setup}
\label{sec: human validation}
We evaluate our annotation framework with two human protocols over the same sampled items.
In \textbf{human verification}, annotators see the framework's output and reasoning and judge whether the output is correct.
In \textbf{independent annotation}, annotators receive the same inputs and task instructions but assign the label themselves without seeing the framework's answer.
Verification asks the narrower question of whether our framework output is defensible under the instructions, allowing annotators to accept edge cases for which more than one answer is plausible; we revisit such cases in Appendix~\ref{app:examples-error-analysis}.

For each of the 11 underlying annotation modules,
we sample 200 rows from the corpus.
Where applicable, we balance the framework label (100 positive and 100 negative) so that the negative class is not swamped by the prevalent positive class.
One hundred outcome-positive rows are shared across the seven taxonomic modules and the two auxiliary attributes, giving an evaluation subset with consistent coverage; the conditionally invoked \gettargetsent module is sampled separately.
The full sampling procedure is detailed in Appendix~\ref{app:human-validation-sampling}.

Each item under each protocol is examined by three annotators with at least undergraduate-level background in computer science. 
Annotators received approximately thirty minutes of training per classifier and were paid \$18.28 per hour for both training and annotation. See Appendix~\ref{app:human-annotators} for details.

\begin{table}[t]
    \centering
    \footnotesize
    {\setlength{\tabcolsep}{2pt}%
        \resizebox{\columnwidth}{!}{%
        \begin{tabular}{@{}lcccc@{}}
            \toprule
            & \multicolumn{2}{c}{\textbf{Verification}}
            & \multicolumn{2}{c}{\textbf{Independent}} \\
            \cmidrule(lr){2-3}\cmidrule(l){4-5}
            \textbf{Classifier} & \textbf{Acc} & \textbf{$\alpha$}
            & \textbf{Acc} & \textbf{$\alpha$} \\
            \midrule
            \multicolumn{5}{l}{\textit{Is it about outcomes?} (\S\ref{subsec:isoutcome})} \\
            \quad\quad\aboutoutcome
                & 0.990 & 0.853 & 0.870 & 0.619 \\
            \multicolumn{5}{l}{\textit{Is it generalised?} (\S\ref{subsec:isgen})} \\
            \quad\quad\isgen
                & 0.990 & 0.760 & 0.765 & 0.065 \\
            \multicolumn{5}{l}{\textit{Is it framed?} (\S\ref{subsec:isframe})} \\
            \quad\quad\refstudy
                & 0.975 & 0.853 & 0.905 & 0.713 \\
            \quad\quad\refstudypop
                & 0.980 & 0.800 & 0.765 & 0.428 \\
            \multicolumn{5}{l}{\textit{Is it generic?} (\S\ref{subsec:isgeneric})} \\
            \quad\quad\ispresenttense
                & 0.985 & 0.933 & 0.980 & 0.933 \\
            \quad\quad\containsquant
                & 0.985 & 0.900 & 0.880 & 0.670 \\
            \quad\quad\removableframe\footnotemark
                & 0.970 & 0.967 & 0.870 & 0.640 \\
            \multicolumn{5}{l}{\textit{Auxiliary attributes} (\S\ref{subsec:flags})} \\
            \quad\quad\containshedging
                & 0.995 & 0.833 & 0.875 & 0.558 \\
            \quad\quad\containsvague
                & 0.975 & 0.771 & 0.710 & 0.090 \\
            \bottomrule
        \end{tabular}}}
    \caption{\textbf{Acc} is majority-to-framework agreement: under
    verification, it is the proportion of items for which the majority judged
    the framework output correct; under independent annotation, it is the
    proportion for which the majority-assigned label matches the framework.
    We also report nominal Krippendorff's $\boldsymbol{\alpha}$.
    Verification $\alpha$ uses reconstructed verifier-implied labels;
    independent $\alpha$ uses directly assigned labels. See
    Appendix~\ref{app:independent-annotation} for details.}
    \label{tab:human-validation}
\end{table}
\footnotetext{For \removableframe, $\alpha$ measures only the Boolean gate
derived from \isreporting and \containsembedded; unlike \textbf{Acc}, it does
not include target-sentence extraction quality. See
Appendix~\ref{app:removable-framing-validation}.}

\paragraph{Agreement and accuracy.}
Table~\ref{tab:human-validation} summarises the results.
We use \textbf{Acc} for
agreement between the human majority vote and our framework.
Under the verification setting, this is the proportion of items for which most annotators judged the framework output correct.
Under independent annotation, it is the proportion for which the majority of directly assigned labels matches the framework output.

For verification, we reconstruct the label implied by each annotator: an accepted framework label is retained and a rejected Boolean label is negated.
We compute Krippendorff's $\alpha$~\citep{krippendorff1989content} over these individual reconstructed labels.
For independent annotation, we compute $\alpha$ directly over the individual labels assigned without seeing the framework output.

Under verification, accuracy is at least $0.970$ for every classifier and $\alpha$ ranges from $0.760$ to $0.967$.
Independent annotation produces lower accuracy ($0.710$--$0.980$), while its $\alpha$ varies substantially across modules ($0.065$--$0.933$).
In particular, agreement is very low for \isgen ($\alpha=0.065$) and \containsvague ($\alpha=0.090$), moderate for \refstudypop ($\alpha=0.428$), and high for \ispresenttense ($\alpha=0.933$).
Thus, the independent annotations do not provide an equally stable reference for every semantic classifier.
We therefore use verification as our primary reliability result and treat independent annotation as a diagnostic.
The verification scores measure whether the framework outputs are defensible under our taxonomy, showing that the LLM annotations are reliable.
Note that the LLM annotations are obtained using a single off-the-shelf open-weight model and do not represent an upper bound on framework performance.
See Appendix~\ref{app:validation_details} and Appendix~\ref{app:independent-annotation} for more details.

\paragraph{Per-label reliability.}
The verification accuracy captures the reliability of a single classifier, but the final taxonomic label for a sentence is determined through a decision tree that combines multiple classifiers (see Fig.~\ref{fig:generalised-decision-tree}).
We therefore translate the per-classifier validation numbers into a per-label reliability score: an estimate of the probability that an LLM annotation carrying a given final label is correct, composed from per-classifier 
accuracy scores.

Let $L$ denote the final label from our taxonomy.
We define the \textbf{data-weighted reliability} for label $L$ by summing over paths $p \in \mathcal{P}_L$, the set of paths through the decision tree that terminate at label $L$
\begin{equation}
    R(L) \;=\; \sum_{p \in \mathcal{P}_L} \frac{n_{p,L}}{n_L} P(L\ \text{is correct}|p)
    \label{eq:label-trust}
\end{equation}
where $n_L$ is the number of claims in the data with final label $L$ and $n_{L,p}$ is the number of those claims that were produced with path $p$.
We can compute the conditional probability as the path reliability
\begin{equation}
    P(L\ \text{is correct}|p) \;=\; \prod_{c \in p} P(v_c\ \text{is correct})
    \label{eq:path-trust}
\end{equation}
where $c$ is a classifier and $v_c$ is the value predicted by that classifier. The individual per-classifier probabilities are the per-classifier accuracies computed from the verification judgements.
Table~\ref{tab:label-dist} shows the per-label reliability scores (see Appendix~\ref{app: trust-score-detail} for a detailed breakdown).
Every final label is therefore accompanied by a quantitative reliability estimate, which downstream analyses can use to weight or filter the annotated data.

\section{Analysis}
\label{sec:analysis}
Using our \datasetname corpus, we conduct five analyses into how generalisations are used in NLP research papers.
Note that the analysis is limited to the sentences annotated as \aboutoutcome (the first filtering step for genericity annotation).
See Appendix \ref{appendix:about_outcome_per_section} for the distribution of \aboutoutcome sentences per section across venues.

\subsection{Trends Across Sections}

\begin{figure*}[!htb]
\centering
\includegraphics[width=0.90\textwidth]{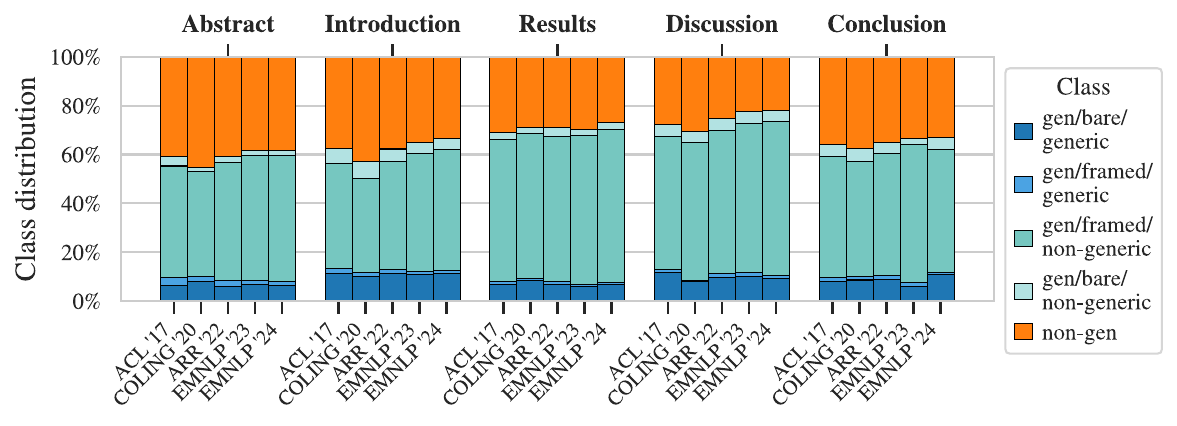}
\vspace{-1em}
\caption{Class distribution of \aboutoutcome sentences across all venues, separated by section and venue.}
\label{fig:gen_class_across_sections}
\end{figure*}

We probe at whether the distribution of generalisations changes per paper section.
Across different venues, the most abundant class was \genfng (Fig.~\ref{fig:gen_class_across_sections}).
Interestingly, this class seems to grow slightly over the years, but further fine-grained analysis is needed to determine why.
The overall ratio of \genast to \ngen sentences is around 60:40, with the most generalisations found in the discussion.
We observe the most \generic claims in the introduction and the fewest in the abstract and results sections. The majority of these are not framed \genolong.

\subsection{Trends Across Tracks}

\begin{figure*}[!htb]
\centering
\includegraphics[width=0.90\textwidth]{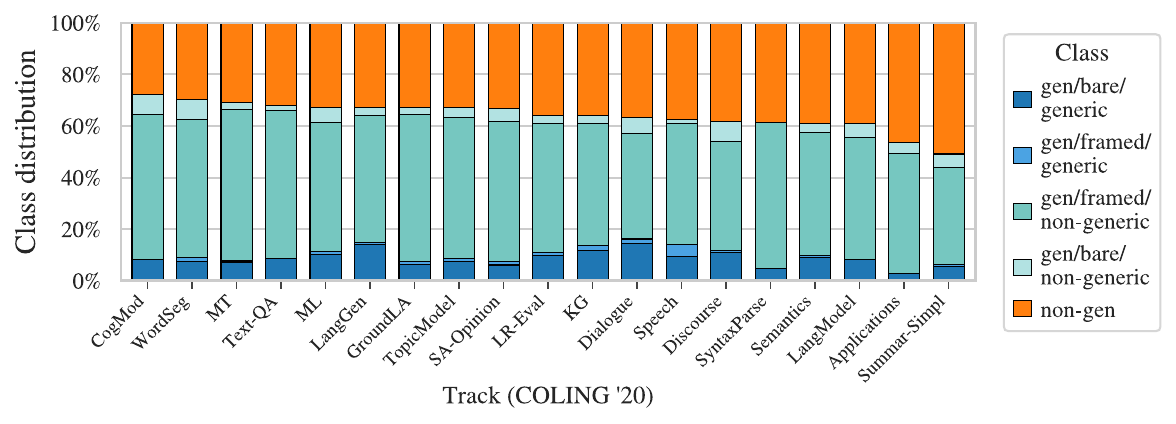}
\vspace{-1em}
\caption{Class distribution of \aboutoutcome sentences in COLING '20 papers, separated by track.}
\label{fig:substantiator_gen_categories_COLING-20}
\end{figure*}

We next probe at whether different sub-topics of NLP research result in different rates of generalisations.
Out of the 5 venues represented in the dataset, only the COLING-2020 papers (89) contained track information.
See Appendix \ref{appendix:tracks} for experiments with assigning soft track labels to other venues.
We visualise the per-track distributions of gen labels for COLING-2020 papers in Figure \ref{fig:substantiator_gen_categories_COLING-20}.

The Applications and Summar-Simpl (Summarization and Simplification) tracks contain the most \ngen sentences.
The Applications class also has one of of the lowest rates of bare generics overall.
This makes intuitive sense - Application papers are usually constrained to a specific subfield (e.g., BioNLP, Social Media), limiting the usage of broad statements.
We also see a large difference in the rates of \ngen across the tracks, with almost half of the sentences in the Summar-Simpl track being non-gen while only ~15\% of the CogMod (Cognitive Modelling) track.
We leave a thorough investigation for the reasons behind track-level generalisation differences for future work.

\subsection{Generalisations and Citation Count}

We next probe at the correlation between a paper's impact (approximated through citation count) and its usage of generalising language.
We retrieve citation counts using SemanticScholar (see Appendix \ref{appendix:citations} for details).
For each paper, we 
compute the distribution for each category over all \aboutoutcome sentences.
In addition to the five categories in our taxonomy, we consider three special groupings: total generalisations (all \genast categories), total generics (both \generic categories), total framed generalisations (both \framed categories).
We then calculate the Spearman rho ($\rho$) correlation between these eight types
(five original labels and three additional) and the total citation count of the paper.
We perform this analysis per venue.

\begin{table}[!t]
\centering
\small
\begin{tabular}{lcc}
\toprule
\textbf{Category} & \textbf{$\rho$} & p-value \\ \midrule
\ngen & -0.15 & 0.173 \\
\genfng & \textbf{0.29} & \textbf{0.005} \\
\geng & \textbf{-0.29} & \textbf{0.007} \\
\geno & -0.06 & 0.604 \\
\genfg & -0.08 & 0.484 \\ \midrule
\generic & \textbf{-0.28} & \textbf{0.008} \\
\genast & 0.15 & 0.176 \\
\genfast & \textbf{0.28} & \textbf{0.008} \\ \bottomrule
\end{tabular}
\caption{Spearman $\rho$ correlation between the generalisation classes and citation counts for ACL '17. Statistically significant effects (p-value < 0.05) in bold.}
\label{tab:spearman_rho_acl17}
\end{table}

Predicting citation counts is a hard task, especially when using a single linguistic feature (generalisability) without \textit{any} further information about semantic content.
Nevertheless, we observe a statistically signification correlation for ACL '17 (Table \ref{tab:spearman_rho_acl17}),  with Spearman $\rho$ of almost 0.3 for \genfng sentences.
This
relationship is similar to that observed by \citet{peters2024hasty_medical}.
Intuitively, stronger papers effectively generalise over their findings,
framing and scoping their claims when necessary to prevent over-claiming.
We find similar (but not significant) trends across other venues (see Appendix \ref{appendix:citations}), indicating this topic requires further study.

\subsection{Hedging tracks generalisation}
\begin{table}[t]
    \centering
    \scriptsize
    \setlength{\tabcolsep}{2pt}
    \begin{tabularx}{\columnwidth}{@{}>{\raggedright\arraybackslash}Xrrrr@{}}
        \toprule
        \textbf{Category} & \textbf{\% hedged} & \textbf{\% of hedged}
            & \textbf{\% vague} & \textbf{\% of vague} \\
        \midrule
        \ngen   & 14.6 & 15.1 & 64.8 & 27.2 \\
        \genfng & 36.7 & 65.5 & 80.5 & 58.3 \\
        \geng   & 40.2 & 11.5 & 80.2 & 9.3 \\
        \geno   & 49.2 & 6.2  & 76.7 & 3.9 \\
        \genfg  & 41.5 & 1.6  & 80.7 & 1.3 \\
        \midrule
        \generic & 40.3 & 13.1 & 80.2 & 10.6 \\
        \genast  & 37.9 & 84.9 & 80.2 & 72.8 \\
        \genfast & 36.8 & 67.1 & 80.5 & 59.5 \\
        \bottomrule
    \end{tabularx}
    \caption{Hedging and vague-descriptor prevalence and composition for all 180K outcome sentences from five venues. The five taxonomy labels partition the corpus; aggregate categories overlap.}
    \label{tab:auxiliary-attribute-analysis}
\end{table}

We next probe whether the use of hedging is associated with generalised claims about outcomes.
Table~\ref{tab:auxiliary-attribute-analysis} reports hedging prevalence and composition for the five taxonomy labels and the three aggregate categories.
Across all 180K fully labelled \aboutoutcome sentences in our corpus,
hedging is
roughly 2.6 times more frequent in generalised than in non-generalised sentences (37.9\% vs.\ 14.6\%).
The inverse direction is even sharper: of all hedged claims about outcomes, around 84.9\% are also annotated as generalised, indicating that hedging is a strong (though not perfect) surface cue for generalisation.

The per-label breakdown refines this picture.
Among generalised labels, the highest hedging rate is on \geno at 49.2\%, while \genfng has the lowest at 36.7\%.
That is, when the generalising frame is made explicit, authors hedge less; bare generalisations that lack a generic noun phrase attract the most hedging, plausibly because the epistemic load is otherwise unmarked.

\subsection{Vague descriptors track generalisation}

Finally, we examine \containsvague, the auxiliary attribute for under-specified adjectives and adverbs introduced in \S\ref{subsec:flags}.
It is
labelled independently from the classifiers
that determine the generalisation label.
Across all five venues, $72.8\%$ of outcome sentences containing a vague descriptor are labelled as generalised; the standard deviation across the five venue-specific proportions is 1.4 percentage points.

Table~\ref{tab:auxiliary-attribute-analysis} shows the association in both directions.
Vague descriptors occur in $80.2\%$ of generalised sentences, compared with $64.8\%$ of non-generalised sentences.
The odds that a generalised sentence contains a vague descriptor are therefore 2.20 times those for a non-generalised sentence.
The association is similar across the five venues.

We treat this co-occurrence as exploratory.
Independent majority agreement with the framework is only $0.710$ for \containsvague, with low chance-corrected agreement ($\alpha=0.090$; Table~\ref{tab:human-validation}).
Moreover, the vague-descriptor and generalisation labels are produced by separate prompts but the same LLM pipeline, so shared annotation artefacts may contribute to the measured association.

\section{Conclusion}

We present a novel taxonomy \taxonomyname for classifying the generality of sentences communicating scientific findings in NLP research.
We also provide a fully automated annotation framework \frameworkname that operationalises the classification of a sentence into one of our five generalisation categories using an ensemble of linguistics-inspired classifiers.
To our knowledge, \frameworkname is the first fully automated solution leveraging instruction-tuned LLMs for annotating generalisation in language, as prior works have either been manual or focused on smaller scale experiments.
We use our \frameworkname to also construct \datasetname, a dataset of 180K NLP claims annotated for generalisation according to our taxonomy \taxonomyname, as well as for hedging and vague descriptors.
Again, to our knowledge, this is the first large-scale dataset annotating generalisation in scientific claims.
Our analyses on \datasetname reveal preliminary patterns that warrant further investigation: the relative percentage of generalisations has grown over time, NLP sub-domains have slightly different distributions, the relative rate of generalisations may be correlated with a paper's citation count, and both hedging and vague descriptors are associated with generalised claims.

\section*{Limitations}
\label{sec:limitations}
Our taxonomy is developed and validated specifically for sentences describing outcomes in NLP papers. While this covers a large portion of claims in a paper, there can also be claims about the work of other authors. These background claims~\citep{lauscher-etal-2018-investigating} are not captured by our taxonomy.
Additionally, the semantics of generics are highly nuanced~\citep[e.g.,][]{krifka1995genericity}. Combining this with the complexity of many scientific sentences means that some edge cases may not be accounted for by our taxonomy. This includes nested claims, which may have components from different categories, and pronoun ambiguity.  

The use of computer graphics papers during initial development~\citep{fisas-etal-2015-discoursive,fisas-etal-2016-multi}, together with the cross-domain foundations discussed in \S\ref{sec: theory}, suggests potential applicability to other empirical scientific domains.
This does not establish cross-domain validity; differences in terminology and discourse conventions, particularly in less constrained genres (e.g., broad social-science discourse), may require adaptation and separate validation.

\frameworkname uses LLM prompting for annotation. While we validate the quality of this framework, the nature of LLMs limits its explainability, reproducibility, and efficiency. Our rule-based baseline (Appendix~\ref{app:rule-baseline}) approaches \frameworkname on several surface-oriented decisions, but substantial performance gaps remain for tasks requiring semantic judgement and for end-to-end classification. We therefore use LLM-based classification consistently across all experiments, while future work could explore hybrid frameworks that replace suitable modules with deterministic rules to improve reproducibility and reduce computational cost.

\section*{Acknowledgements}
We would like to thank the anonymous reviewers for their valuable suggestions.
The authors also thank Huawei Technologies Co. for their generous support towards this research.

\bibliography{custom}

\appendix

\section{LLM Annotation Details}
\label{app:implementation}

\paragraph{Sentence extraction and coverage check.}
Each NLPeer paper is supplied as a structured ITG JSON file that enumerates headings, paragraphs, list items, and formulas in reading order.
We segment every paragraph into sentences with spaCy\footnote{\url{https://spacy.io/}} and retain only those drawn from sections whose heading matches one of the keywords \emph{abstract}, \emph{introduction}, \emph{result}, \emph{discussion}, \emph{conclusion}, or \emph{ablation}. To guard against silent drops from section detection or sentence segmentation, we run a per-paper coverage check on the Introduction: every non-empty token of the raw Introduction text must reappear, after punctuation-and-whitespace normalisation, within the union of the extracted sentences. Papers that fail this check are quarantined and excluded from downstream annotation; the remaining papers proceed with a CSV listing the selected sentences, keyed by (paper, section, within-paper sentence index), which serves as the input to the annotation pipeline described above.

\paragraph{Full decision tree and data flow.}
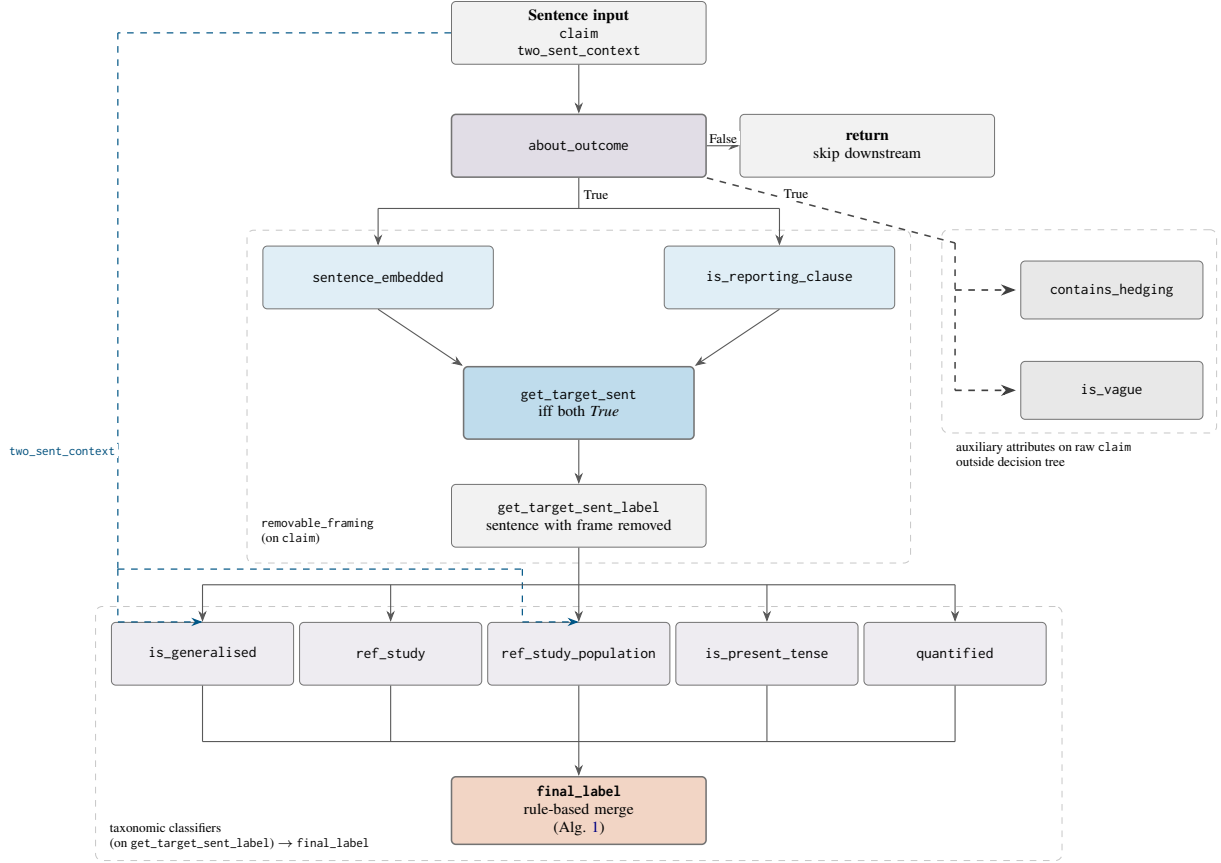
\begin{figure*}[t]
    \centering
    \resizebox{\linewidth}{!}{%
        \begin{tikzpicture}[
            font=\scriptsize,
            >=Stealth,
            io/.style={
                    rectangle, draw=black!55, rounded corners=2pt,
                    fill=gray!10, text width=10em, align=center,
                    inner sep=3pt, minimum height=2.6em
                },
            gate/.style={
                    rectangle, draw=black!55, rounded corners=2pt,
                    fill=MyColour!18, text width=10em, align=center,
                    inner sep=3pt, minimum height=2.6em, line width=0.7pt
                },
            struct/.style={
                    rectangle, draw=black!55, rounded corners=2pt,
                    fill=genBlue!12, text width=9em, align=center,
                    inner sep=3pt, minimum height=2.6em
                },
            extract/.style={
                    rectangle, draw=black!55, rounded corners=2pt,
                    fill=genBlue!25, text width=9em, align=center,
                    inner sep=3pt, minimum height=3.0em, line width=0.7pt
                },
            taxon/.style={
                    rectangle, draw=black!55, rounded corners=2pt,
                    fill=MyColour!10, text width=7em, align=center,
                    inner sep=3pt, minimum height=2.6em
                },
            flag/.style={
                    rectangle, draw=black!55, rounded corners=2pt,
                    fill=gray!18, text width=7em, align=center,
                    inner sep=3pt, minimum height=2.4em
                },
            fin/.style={
                    rectangle, draw=black!55, rounded corners=2pt,
                    fill=GenericFramedColour!35, text width=10em, align=center,
                    inner sep=3pt, minimum height=2.8em, line width=0.7pt
                },
            arr/.style={-{Stealth[length=1.8mm]}, draw=black!65, line width=0.5pt},
            bus/.style={draw=black!65, line width=0.5pt},
            ctxarr/.style={-{Stealth[length=1.8mm]}, dashed, draw=genBlue!75!black, line width=0.55pt},
            ctxbus/.style={dashed, draw=genBlue!75!black, line width=0.55pt},
            flagarr/.style={-{Stealth[length=2.4mm,width=1.8mm]}, dashed, draw=black!75, line width=0.75pt, shorten >=2pt},
            flagbus/.style={dashed, draw=black!72, line width=0.7pt},
            lbl/.style={font=\tiny, inner sep=1pt, fill=white},
            ctxlbl/.style={font=\tiny, inner sep=1pt, fill=white, text=genBlue!70!black},
            boxlbl/.style={font=\tiny, inner sep=1pt, fill=white, align=left}
            ]

            \node[io]      (in)    at ( 0,    0.0) {\textbf{Sentence input}\\ \texttt{claim}\\ \texttt{two\_sent\_context}};
            \node[gate]    (ao)    at ( 0,   -1.8) {\aboutoutcome};
            \node[io]      (stop)  at ( 4.6, -1.8) {\textbf{return}\\ skip downstream};

            \node[struct]  (emb)   at (-3.2, -3.9) {\containsembedded};
            \node[struct]  (rep)   at ( 3.2, -3.9) {\isreporting};

            \node[extract] (gts)   at ( 0,   -5.9) {\gettargetsent\\ \text{iff both} \textit{True}};
            \node[io]      (tgt)   at ( 0,   -7.7) {\texttt{get\_target\_sent\_label}\\ \text{sentence with frame removed}};

            \node[taxon]   (gen)   at (-6.0, -9.9) {\isgen};
            \node[taxon]   (rs)    at (-3.0, -9.9) {\refstudy};
            \node[taxon]   (rsp)   at ( 0.0, -9.9) {\refstudypop};
            \node[taxon]   (tense) at ( 3.0, -9.9) {\ispresenttense};
            \node[taxon]   (quant) at ( 6.0, -9.9) {\containsquant};

            \node[fin]     (final) at ( 0,  -12.4) {\textbf{\texttt{final\_label}}\\ rule-based merge\\ (Alg.~\ref{alg:ann-decision-tree})};

            \node[flag]    (hed)   at ( 8.5, -4.1) {\containshedging};
            \node[flag]    (vague) at ( 8.5, -5.7) {\containsvague};

            \draw[arr] (in) -- (ao);
            \draw[arr] (ao.east) -- node[lbl, above] {False} (stop.west);

            \coordinate (aoT) at (0, -2.8);
            \draw[bus] (ao.south) -- (aoT) node[lbl, right=0.1em, pos=0.55] {True};
            \draw[bus] (-3.2, -2.8) -- (3.2, -2.8);
            \draw[arr] (-3.2, -2.8) -- (emb.north);
            \draw[arr] ( 3.2, -2.8) -- (rep.north);

            \draw[arr] (emb.south) -- (gts.north west);
            \draw[arr] (rep.south) -- (gts.north east);

            \draw[arr] (gts) -- (tgt);

            \coordinate (ctxLeftTop) at (-7.35, 0.0);
            \coordinate (ctxBusLeft) at (-7.35, -8.55);
            \coordinate (ctxBusRight) at (-0.9, -8.55);
            \draw[ctxbus] (in.west) -- (ctxLeftTop);
            \draw[ctxbus] (ctxLeftTop) -- node[ctxlbl, left, pos=0.78] {\texttt{two\_sent\_context}} (ctxBusLeft);
            \draw[ctxbus] (ctxBusLeft) -- (ctxBusRight);
            \draw[ctxarr] (ctxBusLeft) |-(gen.north);
            \draw[ctxarr] (ctxBusRight) |- (rsp.north);

            \coordinate (busT) at (0, -8.8);
            \draw[bus] (tgt.south) -- (busT);
            \draw[bus] (-6.0, -8.8) -- (6.0, -8.8);
            \draw[arr] (-6.0, -8.8) -- (gen.north);
            \draw[arr] (-3.0, -8.8) -- (rs.north);
            \draw[arr] ( 0.0, -8.8) -- (rsp.north);
            \draw[arr] ( 3.0, -8.8) -- (tense.north);
            \draw[arr] ( 6.0, -8.8) -- (quant.north);

            \coordinate (busF) at (0, -11.3);
            \draw[bus] (gen.south)   -- (-6.0, -11.3);
            \draw[bus] (rs.south)    -- (-3.0, -11.3);
            \draw[bus] (rsp.south)   -- ( 0.0, -11.3);
            \draw[bus] (tense.south) -- ( 3.0, -11.3);
            \draw[bus] (quant.south) -- ( 6.0, -11.3);
            \draw[bus] (-6.0, -11.3) -- (6.0, -11.3);
            \draw[arr] (busF) -- (final.north);

            \coordinate (flagJoin) at (6.0, -3.35);
            \coordinate (flagStemTop) at (6.0, -4.1);
            \coordinate (flagStemBottom) at (6.0, -5.7);
            \draw[flagbus] (ao.south east) -- node[lbl, above, pos=0.36] {True} (flagJoin);
            \draw[flagbus] (flagJoin) -- (flagStemBottom);
            \draw[flagarr] (flagStemTop) -- (hed.west);
            \draw[flagarr] (flagStemBottom) -- (vague.west);

            \begin{scope}[on background layer]
                \node[fit=(emb)(rep)(gts)(tgt), draw=black!22, rounded corners=3pt,
                    inner sep=7pt, dashed, line width=0.35pt,
                ] (structbox) {};
                \node[fit=(gen)(quant)(final), draw=black!22, rounded corners=3pt,
                    inner sep=7pt, dashed, line width=0.35pt,
                ] (taxbox) {};
                \node[fit=(flagJoin)(flagStemTop)(flagStemBottom)(hed)(vague), draw=black!22, rounded corners=3pt,
                    inner sep=6pt, dashed, line width=0.35pt,
                ] (flagbox) {};
            \end{scope}
            \node[boxlbl, anchor=north west] at ([xshift=0.5em,yshift=2em]structbox.south west)
            {\removableframe\\(on \texttt{claim})};
            \node[boxlbl, anchor=south west] at ([xshift=0.5em,yshift=0.35em]taxbox.south west)
            {taxonomic classifiers\\(on \texttt{get\_target\_sent\_label}) $\rightarrow$ \texttt{final\_label}};
            \node[boxlbl, anchor=north west] at ([xshift=0.5em,yshift=-0.25em]flagbox.south west)
            {auxiliary attributes on raw \texttt{claim}\\outside decision tree};
        \end{tikzpicture}%
    }
    \caption{%
        Data flow across the annotation pipeline (one sentence at a time).
        The dashed right branch records auxiliary attributes that do not feed \texttt{final\_label}.
    }
    \label{fig:ann-prompt-data-flow}
\end{figure*}

\begin{algorithm}[t]
    \caption{Get final taxonomy label}
    \label{alg:ann-decision-tree}
    \begin{algorithmic}[1]
        \footnotesize
        \Require Boolean labels from the seven taxonomic classifiers (\S\ref{app:human-validation-sampling}), given the \aboutoutcome{} label is \textsc{true}
        \Ensure Final label $\ell$
        \If{\isgen\ is \textsc{false}}
        \State $\ell \gets$ \ngen
        \Else
        \If{\refstudypop\ is \textsc{true}}
        \State $\ell \gets$ \genfng
        \ElsIf{\refstudy\ is \textsc{true}}
        \If{\ispresenttense\ is \textsc{false}}
        \State $\ell \gets$ \genfng
        \ElsIf{\isreporting\ is \textsc{false}}
        \State $\ell \gets$ \genfng
        \ElsIf{\containsembedded\ is \textsc{false}}
        \State $\ell \gets$ \genfng
        \ElsIf{\containsquant\ is \textsc{true}}
        \State $\ell \gets$ \genfng
        \Else
        \State $\ell \gets$ \genfg
        \EndIf
        \Else
        \If{\ispresenttense\ is \textsc{false}}
        \State $\ell \gets$ \geno
        \ElsIf{\containsquant\ is \textsc{true}}
        \State $\ell \gets$ \geno
        \Else
        \State $\ell \gets$ \geng
        \EndIf
        \EndIf
        \EndIf
        \State \Return $\ell$
    \end{algorithmic}
\end{algorithm}

A more fine-grained decision tree is shown in Algorithm~\ref{alg:ann-decision-tree}.

Figure~\ref{fig:ann-prompt-data-flow} summarises how classifier outputs are chained before the final label is computed.
\aboutoutcome gates the rest: when False, no other LLM call is made for the sentence.
\gettargetsent only calls the LLM when both \isreporting and \containsembedded are True; otherwise \texttt{get\_target\_sent\_label} is set to the original \texttt{claim}.
The five taxonomic classifiers all read \texttt{get\_target\_sent\_label}; the blue dashed path marks the additional \texttt{two\_sent\_context} input used by \isgen and \refstudypop.
These five classifiers feed the rule-based \texttt{final\_label} of Algorithm~\ref{alg:ann-decision-tree}.
\containshedging and \containsvague run on the raw \texttt{claim} and sit outside the decision tree.

\paragraph{Hardware and performance.}
We ran experiments on a mixture of NVIDIA A100 (80\,GB), H100 (80\,GB), and H200 (80\,GB) GPUs, serving gpt-oss-120b \citep{agarwal2025gpt} via vLLM \citep{kwon2023efficient} with asynchronous requests.
As a rough throughput estimate, annotating a 100-sentence paper with $\text{num\_passes}{=}1$ on a 4$\times$H200 node takes approximately 10 minutes, though this figure varies with the fraction of sentences that pass the \aboutoutcome gate and with the inherent variability of asynchronous, parallelised inference.

\paragraph{Annotation statistics.}
As introduced in Section~\ref{sec:dataset}, each \aboutoutcome sentence is annotated three times by the LLM, and the categorical classifier outputs are merged by majority vote across the three passes, with pass 1 as the tie-breaker.
Across the ten Boolean modules reported in the paper, this produces $2{,}019{,}550$ majority-voted classifier cells (gated by \aboutoutcome for downstream classifiers).
Only $13$ remain missing and $94$ are tied after the merge: $11$ sentences across $7$ papers are incomplete, and a further $94$ sentences across $94$ papers carry a tie-broken label.
Missingness is already low before merging ($830$ empty cells summed across the three passes, $\sim 0.01\%$), so the vote serves chiefly to suppress single-pass failures rather than to compensate for systematic gaps.
We conclude that the annotated corpus is effectively complete at scale.

\section{Human Verification Details}
\label{app:validation_details}

\paragraph{Human verification.}
\label{app:human-annotators}
We recruited seven annotators with at least undergraduate-level training in computer science and prior experience providing teaching support. All annotators were familiar with the NLP field and with scientific writing style.

In this verification protocol, annotators did not relabel sentences from scratch.
Instead, they audited the LLM pipeline: for each classifier, they judged whether the model's decision was correct, given the same inputs available to the model (\S~\ref{sec: human validation}). 
Each sampled row was judged independently by three annotators. Assignment sheets for each annotation step were identical across annotators, ensuring that every row received three votes.

Before annotation, annotators received approximately $30$ minutes of training for each annotation step. During training, they reviewed the corresponding prompt template (Appendix~\ref{app:prompt_templates}) and were instructed to judge each LLM label strictly according to that template. The workbooks themselves contained only the instantiated inputs and outputs for each sampled row.

For each classifier, we distributed one Excel workbook per annotator, with one file per classifier. Each workbook contained the following columns:
\begin{itemize}
    \item Row identifiers: corpus split, paper path, and within-paper sentence index.
    \item Sentence context matching the prompt placeholders for that classifier: typically the original sentence \texttt{\{claim\}} or the extracted framed sentence \texttt{\{get\_target\_sent\_label\}}, sometimes supplemented with a two-sentence local context \texttt{\{two\_sent\_context\}}.
    \item The LLM label and its free-text reasoning.
\end{itemize}

The annotators were asked to judge each LLM label strictly according to the corresponding prompt template, and fill in the following fields:
\begin{itemize}
    \item A mandatory True/False field recording whether the LLM decision is correct.
    \item An optional free-text comments field.
\end{itemize}

Annotators were paid equivalent to US\$18.28 per hour for both training and annotation.
They are informed that the annotation task is part of a research project and that their contributions will be used to evaluate the performance of a annotation framework, and the data will be made available to the public.
The data collection protocol was reviewed and approved by our institution's research ethics review board prior to recruitment.

\paragraph{Human verification set sampling.}
\label{app:human-validation-sampling}

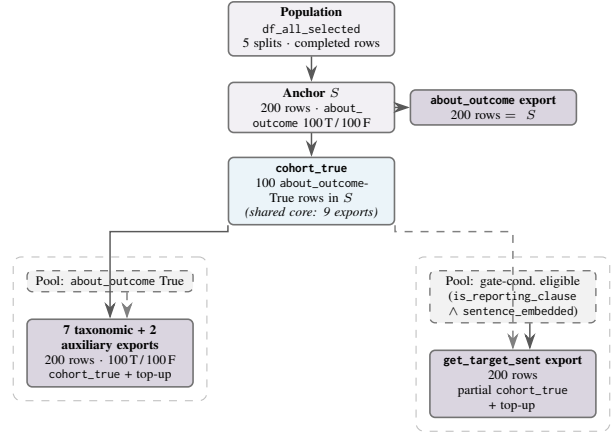
\begin{figure}[t]
    \centering
    \begin{tikzpicture}[
            scale=0.7,
            transform shape,
            font=\scriptsize,
            >=Stealth,
            block/.style={
                    rectangle,
                    draw=black!55,
                    rounded corners=2pt,
                    text width=7.6em,
                    align=center,
                    inner sep=3pt
                },
            stage/.style={block, fill=MyColour!8, line width=0.6pt},
            cohortbox/.style={block, fill=genBlue!8, line width=0.6pt},
            poolbox/.style={block, fill=gray!10, line width=0.5pt, dashed},
            exportbox/.style={block, fill=MyColour!14, line width=0.6pt},
            termbox/.style={exportbox, fill=MyColour!22},
            arr/.style={
                    -{Stealth[length=2.2mm, width=1.5mm]},
                    draw=black!65,
                    line width=0.55pt
                },
            arrpool/.style={arr, dashed, black!50}
        ]
        \node[stage] (pop) {%
            \textbf{Population}\\[0.1em]
            \texttt{df\_all\_selected}\\
            5 splits $\cdot$ completed rows
        };
        \node[stage, below=0.48cm of pop] (anchor) {%
            \textbf{Anchor} $S$\\
            200 rows $\cdot$ \aboutoutcome\ 100\,T\,/\,100\,F
        };
        \node[termbox, right=0.28cm of anchor] (ao) {%
            \textbf{\aboutoutcome\ export}\\
            200 rows $= S$
        };

        \node[cohortbox, below=0.45cm of anchor] (cohort) {%
            \textbf{\texttt{cohort\_true}}\\
            100 \aboutoutcome-True rows in $S$\\
            \textit{(shared core: 9 exports)}
        };

        \node[poolbox, below left=0.85cm and 0.65cm of cohort] (poolao) {%
            Pool: \aboutoutcome\ True
        };
        \node[termbox, below=0.50cm of poolao] (sty) {%
            \textbf{7 taxonomic + 2 auxiliary exports}\\
            200 rows $\cdot$ 100\,T\,/\,100\,F\\
            \texttt{cohort\_true} + top-up
        };

        \node[poolbox, below right=0.85cm and 0.65cm of cohort] (poolgt) {%
            Pool: gate-cond.\ eligible\\
            (\isreporting\ $\wedge$\ \containsembedded)
        };
        \node[termbox, below=0.50cm of poolgt] (gt) {%
            \textbf{\gettargetsent\ export}\\
            200 rows \\
            partial \texttt{cohort\_true} + top-up
        };

        \draw[arr] (pop) -- (anchor);
        \draw[arr] (anchor.east) -- (ao.west);
        \draw[arr] (anchor) -- (cohort);

        \draw[arr] (cohort.south west) -- ++(0,-0.12) -| (sty.north);
        \draw[arrpool] ($(poolao.south)+(0.32,0)$) -- ($(sty.north)+(0.32,0)$);

        \draw[arrpool] (cohort.south east) -- ++(0,-0.12) -| (gt.north);
        \draw[arr] ($(poolgt.south)+(0.32,0)$) -- ($(gt.north)+(0.32,0)$);

        \begin{scope}[on background layer]
            \node[fit=(poolao)(sty), draw=black!25, rounded corners=3pt,
                inner sep=5pt, dashed, line width=0.4pt] {};
            \node[fit=(poolgt)(gt), draw=black!25, rounded corners=3pt,
                inner sep=5pt, dashed, line width=0.4pt] {};
        \end{scope}
    \end{tikzpicture}
    \caption{%
        Human-validation sampling pipeline (seed~42).
        Anchor~$S$ stratifies on \aboutoutcome; \texttt{cohort\_true} is the 100-row core reused in the seven taxonomic exports plus \containshedging and \containsvague, but is \emph{not} the direct core for \gettargetsent.
        For \gettargetsent, a dashed arrow shows \texttt{cohort\_true} feeding the module-specific export via conditional filtering; a solid arrow then shows the pool as the primary sampling source.
        For taxonomic exports, a solid arrow shows direct cohort-core inclusion; a dashed arrow shows external top-up from the \aboutoutcome-True eligible pool.
    }
\end{figure}

We draw verification rows from the merged annotated corpus: all sentences with completed LLM annotation across five NLPeer subsets, majority-voted per classifier and keyed by corpus split, paper index, and within-paper sentence index.
For each of the 11 annotation modules we export 200 rows (random seed~42).

Out of the 11 annotation modules, \aboutoutcome is the gating module, \gettargetsent is a helper that provides input for downstream modules, and \containshedging and \containsvague are auxiliary attributes outside the decision tree.
We call the remaining seven modules, which directly determine the final label, taxonomic modules.

We first form an \emph{anchor} set~$S$ by stratifying on merged \aboutoutcome{} labels (100 rows with LLM label True and 100 with False) without balancing by corpus split or paper.
Let \texttt{cohort\_true} denote the 100 rows in~$S$ with \aboutoutcome{} True; this cohort is the \textbf{shared core} included in the seven taxonomic verification sheets and both auxiliary-attribute sheets.

For \gettargetsent, since whether it is invoked is conditional on \isreporting and \containsembedded, we retain eligible rows from \texttt{cohort\_true} and top up from other \aboutoutcome-True rows that satisfy both conditions.
The audited error rate $e_{\text{gts},\text{T}}$ therefore estimates extraction quality on rows where the gating predicates fired; the role of this rate in the per-label reliability score is detailed below in Appendix~\ref{app: trust-score-detail}.

For each of the seven taxonomic modules that implement the generalisation decision tree and the two auxiliary attributes, we build a 200-row export balanced 100/100 on that module's LLM label: sample up to 100 True and 100 False rows from \texttt{cohort\_true} (when available), then top up without replacement from the pool of \aboutoutcome-True sentences until the quota is met.
The \aboutoutcome{} verification set is anchor~$S$ itself.

\paragraph{Per-label reliability: \gettargetsent and the asymmetric veto rule.}
\label{app: trust-score-detail}
The path product $S_p$ in \eqref{eq:path-trust} treats \gettargetsent as one more classifier in the sequence actually evaluated by the pipeline. Because \gettargetsent runs only when both \isreporting and \containsembedded are predicted true, this asymmetric gating warrants a separate accounting of how it can fail. Let $T$ denote the truth-side trigger event ``both \isreporting and \containsembedded are truly true'', and let $L_p$ be the LLM gate decision implied by the path $p$. We decompose the path-level failure into three disjoint events:
\begin{itemize}[itemsep=0em]
    \item $V_{\text{tree}}$: a tree classifier on $p$ is wrong; this is captured by the marginal factors $(1-e_{c,v_c})$ in \eqref{eq:path-trust}.
    \item $V_{\text{ext}}$: \gettargetsent was invoked but the extraction is wrong; captured by $(1-e_{\text{gts},\text{T}})$ on $L_p = (\text{T},\text{T})$ paths.
    \item $V_{\text{skip}}$: $T$ holds but $L_p \ne (\text{T},\text{T})$; the LLM skipped \gettargetsent when it should have run, so downstream classifiers received the original sentence in place of the extracted target.
\end{itemize}
Over-calling \gettargetsent (i.e., $L_p = (\text{T},\text{T})$ but $\neg T$) is not counted as a failure mode: the prompt is designed to return the original sentence in that case, so its only contribution to the path is the extraction pass rate $1 - e_{\text{gts},\text{T}}$.

Marginalising over the truth gates yields a case-wise definition of the gate-block factor $M_p$ of $S_p$, with $r$ and $m$ shorthand for the \isreporting and \containsembedded subscripts:
\begin{align*}
    L_p & = (\text{T},\text{T}):  M_p = (1-e_{r,\text{T}})(1-e_{m,\text{T}})(1-e_{\text{gts},\text{T}}), \\
    L_p & = (\text{T},\text{F}):  M_p = (1-e_{r,\text{T}})(1-e_{m,\text{F}}),                            \\
    L_p & = (\text{F},\,\cdot\,):  M_p = (1-e_{r,\text{F}}).
\end{align*}
On non-$(\text{T},\text{T})$ paths the F-stratum gate factor $(1 - e_{c,\text{F}})$ already equals $\Pr(\neg V_{\text{skip}})$: ``LLM gate F right'' implies ``truth gate F'' implies ``no missed call''. Adding a separate $V_{\text{skip}}$ term would therefore double-count. Hence the path product in \eqref{eq:path-trust}, evaluated on the actually executed sequence, is numerically the marginalised score; the asymmetric rule does not add a new factor, only a new interpretation.

The end-to-end shared-set check in Section~\ref{sec: human validation} reports the two corresponding row-level vetoes separately: a skip-veto fires when the rectified path triggers \gettargetsent but the LLM did not call it, and an extraction-veto fires when the LLM called \gettargetsent and the human majority judged the extraction wrong. Over-calls remain unvetoed and are absorbed into the corpus-level $e_{\text{gts},\text{T}}$.

\paragraph{Agreement and reconstructed labels.}
The verification judgements themselves (LLM correct vs.\ LLM incorrect) are heavily skewed towards the positive class, making chance-corrected agreement on those judgements difficult to interpret.
For each Boolean module, we instead reconstruct the task label implied by each verifier.
If the verifier accepts the framework label, we retain it; if the verifier rejects it, we negate it.
We then compute nominal Krippendorff's $\alpha$ over these individual reconstructed labels before majority voting.
This moves the chance correction onto the task-label scale while preserving the observed pairwise disagreements.
The reconstruction is unavailable for \gettargetsent because rejecting an extracted string does not identify the verifier's preferred alternative.

\section{Independent Annotation Details}
\label{app:independent-annotation}

\begin{table*}[t]
    \centering
    \footnotesize
    {\setlength{\tabcolsep}{3pt}%
        \begin{tabular}{@{}lcccccccc@{}}
            \toprule
            & \multicolumn{4}{c}{\textbf{Verification}}
            & \multicolumn{4}{c}{\textbf{Independent}} \\
            \cmidrule(lr){2-5}\cmidrule(l){6-9}
            \textbf{Classifier}
            & \textbf{Acc} & \textbf{Acc$^+$} & \textbf{Acc$^-$} & \textbf{$\alpha$}
            & \textbf{Acc} & \textbf{Acc$^+$} & \textbf{Acc$^-$} & \textbf{$\alpha$} \\
            \midrule
            \multicolumn{9}{l}{\textit{Is it about outcomes?} (\S\ref{subsec:isoutcome})} \\
            \quad\quad\aboutoutcome
                & 0.990 & 0.990 & 0.990 & 0.853
                & 0.870 & 0.810 & 0.930 & 0.619 \\
            \multicolumn{9}{l}{\textit{Is it generalised?} (\S\ref{subsec:isgen})} \\
            \quad\quad\isgen
                & 0.990 & 0.980 & 1.000 & 0.760
                & 0.765 & 0.590 & 0.940 & 0.065 \\
            \multicolumn{9}{l}{\textit{Is it framed?} (\S\ref{subsec:isframe})} \\
            \quad\quad\refstudy
                & 0.975 & 0.960 & 0.990 & 0.853
                & 0.905 & 0.850 & 0.960 & 0.713 \\
            \quad\quad\refstudypop
                & 0.980 & 0.980 & 0.980 & 0.800
                & 0.765 & 0.770 & 0.760 & 0.428 \\
            \multicolumn{9}{l}{\textit{Is it generic?} (\S\ref{subsec:isgeneric})} \\
            \quad\quad\gettargetsent\footnotemark
                & 1.000 & 1.000 & -- & --
                & 0.865 & 0.865 & -- & -- \\
            \quad\quad\ispresenttense
                & 0.985 & 0.990 & 0.980 & 0.933
                & 0.980 & 0.990 & 0.970 & 0.933 \\
            \quad\quad\containsquant
                & 0.985 & 0.970 & 1.000 & 0.900
                & 0.880 & 0.770 & 0.990 & 0.670 \\
            \quad\quad\isreporting
                & 0.985 & 0.970 & 1.000 & 0.887
                & 0.950 & 0.930 & 0.970 & 0.637 \\
            \quad\quad\containsembedded
                & 0.925 & 0.850 & 1.000 & 0.810
                & 0.890 & 0.780 & 1.000 & 0.415 \\
            \multicolumn{9}{l}{\textit{Auxiliary attributes} (\S\ref{subsec:flags})} \\
            \quad\quad\containshedging
                & 0.995 & 0.990 & 1.000 & 0.833
                & 0.875 & 0.760 & 0.990 & 0.558 \\
            \quad\quad\containsvague
                & 0.975 & 0.950 & 1.000 & 0.771
                & 0.710 & 0.450 & 0.970 & 0.090 \\
            \bottomrule
        \end{tabular}}
    \caption{Per-module results before merging \isreporting, \containsembedded,
    and \gettargetsent into \removableframe. \textbf{Acc} is
    majority-to-framework agreement: under verification, the majority judged
    the framework output correct; under independent annotation, the
    majority-assigned label matches the framework output.
    \textbf{Acc$^+$} and \textbf{Acc$^-$} report this proportion conditional
    on positive and negative framework outputs.
    Verification $\boldsymbol{\alpha}$ uses reconstructed verifier-implied
    labels, while independent $\boldsymbol{\alpha}$ uses directly assigned
    labels.}
    \label{tab:human-validation-unmerged}
\end{table*}
\footnotetext{Label-level $\alpha$ is unavailable for \gettargetsent. Raw pairwise agreement on whether the extraction matches the framework
is $0.937$ under verification and $0.797$ under independent annotation.}

\paragraph{Protocol and sampling.}
Independent annotation complements the human verification described in
Appendix~\ref{app:validation_details}.
We reuse the same 200 sampled items for each of the eleven underlying
annotation modules, but withhold the framework output and its reasoning.
Annotators receive only the task instructions and the text fields supplied to
the corresponding classifier.
For Boolean modules, they assign a True/False label directly; for
\gettargetsent, they write the target proposition.
Each item receives three annotations.

\paragraph{Metrics.}
\textbf{Acc} denotes majority-to-framework agreement in both tables.
Under verification, an item is correct when the majority judged the framework
output correct.
Under independent annotation, a Boolean item is correct when the majority of
the three directly assigned labels matches the framework output.
\textbf{Acc$^+$} and \textbf{Acc$^-$} report this quantity separately for
positive and negative framework outputs.
Krippendorff's $\alpha$ is computed over the three individual labels for each
item, before taking the majority.

The labels entering $\alpha$ differ by protocol.
Let $y_i$ be the framework's Boolean label for item $i$, and let $c_{ia}$ denote
whether verifier $a$ judged it correct.
The verifier-implied label is
\begin{equation}
    \widetilde{y}_{ia} =
    \begin{cases}
        y_i & \text{if } c_{ia}=\text{True}, \\
        \neg y_i & \text{if } c_{ia}=\text{False}.
    \end{cases}
\end{equation}
Verification $\alpha$ is computed over $\widetilde{y}_{ia}$.
Independent $\alpha$ is instead computed directly over each annotator's raw
True/False label.
Both coefficients therefore operate on the same task-label scale and sampled
items.
However, the reconstructed verification labels remain conditioned on seeing
the framework output and reasoning, so they do not constitute independent
annotations.

\paragraph{Per-module results.}
Table~\ref{tab:human-validation} reports the nine modules after consolidating
the three removable-framing operations, while
Table~\ref{tab:human-validation-unmerged} reports all eleven operations.
Independent majority agreement with the framework ranges from $0.710$ to
$0.980$ in the consolidated table.
Verification $\alpha$ ranges from $0.760$ to $0.967$, whereas independent
$\alpha$ ranges from $0.065$ to $0.933$.
The largest protocol difference occurs for \isgen ($0.760$ vs.\ $0.065$).
Independent $\alpha$ is also low for \containsvague ($0.090$), and remains lower for \refstudypop ($0.428$) and
\containsembedded ($0.415$), while \ispresenttense reaches $0.933$ under both
protocols.
Thus, independently applying the instructions is highly reproducible for some
surface-oriented modules but does not yield a stable reference for every
semantic distinction.
The \removableframe coefficients measure its Boolean gate and are derived in
Appendix~\ref{app:removable-framing-validation}.

\paragraph{Target-sentence extraction.}
We normalise each independently written target with Unicode compatibility
normalisation and case folding, then remove punctuation and whitespace before
exact comparison.
All 200 items receive three target strings.
The three normalised strings are identical for 138 items, two distinct strings
occur for 52 items, and three distinct strings occur for 10 items.
The human majority matches the framework extraction for 173 items
($\text{Acc}=0.865$).
We do not report label-level $\alpha$ for this free-text operation because the Boolean label-level $\alpha$ used above is not applicable.
The footnote to Table~\ref{tab:human-validation-unmerged} instead reports raw
pairwise agreement on whether each extraction matches the framework.

\paragraph{End-to-end comparison and interpretation.}
On the 100 items shared across the decision-tree modules, substituting the
independent human-majority decisions into the final tree gives a final label
matching the original framework output for 48 items.
The corresponding verification result is 96 items, and the two human-implied
final labels agree on 61 items.
These differences compound the module-level disagreement and show that
verification and independent annotation answer different questions:
verification assesses whether a proposed output is defensible, whereas
independent annotation measures whether annotators reproduce that output
without seeing it.

The low independent agreement for \isgen and the moderate agreement for
\refstudypop also help contextualise the rule-based baseline in
Appendix~\ref{app:rule-baseline}.
If trained annotators have difficulty reproducing these semantic distinctions
without a proposed answer, deterministic surface rules should not be expected
to recover them reliably.

\section[Composite Human Evaluation]{Composite Label: \removableframe}
\label{app:removable-framing-validation}

Table~\ref{tab:human-validation} consolidates \isreporting, \containsembedded, and \gettargetsent into a single \removableframe row.
This section defines the merge and explains how its extraction-aware accuracy and Boolean-gate Krippendorff's $\alpha$ are computed under both human protocols.

\paragraph{Population.}
We inner-join the \isreporting and \containsembedded human-annotation sheets on the shared row key \texttt{(split, input\_path, row\_idx)}, then intersect with the 100 rows common to all taxonomic evaluation exports.
This yields 100 rows.
The \gettargetsent annotation sheet is left-joined onto the same key set; only 28 of the 100 rows have both \isreporting and \containsembedded labelled True by the framework and therefore have a \gettargetsent human vote.

\paragraph{Merged framework label.}
The composite framework label used to stratify Acc$^+$ / Acc$^-$ is
\begin{equation}
    y_{\text{rf}} = y_{\text{irc}} \wedge y_{\text{ces}},
\end{equation}
where $y_{\text{irc}}$ and $y_{\text{ces}}$ are the framework booleans for \isreporting and \containsembedded respectively.
This mirrors the pipeline gate that determines whether \gettargetsent is invoked: 28 rows have $y_{\text{rf}} = \text{True}$ and 72 have $y_{\text{rf}} = \text{False}$.

\paragraph{Composite majority accuracy.}
For each of the 100 rows, we compute a row-level correctness outcome as follows.
Let $\hat{m}_c$ denote whether the human majority judged framework module $c$'s output correct on that row.
The composite majority verdict is
\begin{equation}
    \begin{aligned}
        \mathrm{correct}_{\text{rf}}
          & = \hat{m}_{\text{irc}} \wedge \hat{m}_{\text{ces}} \wedge g,                                   \\
        g & = \begin{cases}
                  \hat{m}_{\text{gts}} & \text{if } y_{\text{rf}} = \text{True}, \\
                  \text{True}          & \text{otherwise,}
              \end{cases}
    \end{aligned}
\end{equation}
where $\hat{m}_{\text{gts}}$ is the human majority judgement for \gettargetsent on that row. This matches the trust-score path product in \eqref{eq:path-trust}: on non-invoked rows the extraction factor is absent, so the audited pass rate already conditions on the correct event.

\paragraph{Boolean-gate Krippendorff's $\alpha$.}
For verification, let $\widetilde{r}_{ia}$ and $\widetilde{e}_{ia}$ be the
labels implied by verifier $a$ for item $i$ on \isreporting and
\containsembedded, respectively.
As described in Appendix~\ref{app:validation_details}, each label retains the
framework value when the verifier accepts it and negates the value when the
verifier rejects it.
For independent annotation, let $r_{ia}$ and $e_{ia}$ denote the corresponding
raw labels.
We derive one removable-framing gate label per item and annotator:
\begin{equation}
    \begin{aligned}
        \widetilde{m}_{ia}^{\mathrm{ver}}
            &= \widetilde{r}_{ia} \wedge \widetilde{e}_{ia}, \\
        m_{ia}^{\mathrm{ind}}
            &= r_{ia} \wedge e_{ia}.
    \end{aligned}
    \label{eq:rf-alpha-labels}
\end{equation}
We compute nominal Krippendorff's $\alpha$ separately over
$\{\widetilde{m}_{ia}^{\mathrm{ver}}\}$ and
$\{m_{ia}^{\mathrm{ind}}\}$ on the same 100 items, with three labels per item.
The resulting coefficients are $0.967$ for verification and $0.640$ for
independent annotation.

\paragraph{Composite results.}
Under verification setting, 97 of the 100 composite decisions agree with the framework ($\text{Acc}=0.970$).
Under independent annotation, 87 decisions agree: 24 of 28 positive framework outputs ($\text{Acc}^+=0.857$) and 63 of 72 negative outputs ($\text{Acc}^-=0.875$).
Unlike the labels in Equation~\ref{eq:rf-alpha-labels}, these accuracy values
also require \gettargetsent to match when the framework invokes extraction.
The $\alpha$ values for \removableframe in Table~\ref{tab:human-validation} measure only agreement on the Boolean
removable-framing gate; \gettargetsent extraction quality is reported separately in Appendix~\ref{app:independent-annotation}.
Table~\ref{tab:human-validation-unmerged} reports both protocols for the eleven individual modules before \isreporting, \containsembedded, and \gettargetsent are merged.

\section{Rule-based Baseline}
\label{app:rule-baseline}

To explore how much of \frameworkname can be reproduced without an LLM, we implement deterministic counterparts for nine annotation modules that are used in the final decision tree (Algorithm~\ref{alg:ann-decision-tree}) for determining the generalisation label.
Each counterpart translates the corresponding prompt instructions into regular-expression and lexicon rules, supplemented by spaCy part-of-speech, morphological, and dependency features.
The baseline receives the same text fields as \frameworkname, executes the modules in the same dependency order, and applies the same final decision tree.
The rules were derived from the prompts and frozen before evaluation; they do not use LLM predictions, rationales, or human annotations.
We report the three operations consolidated as \removableframe in the main paper separately so that target-sentence extraction can be evaluated directly.

\paragraph{\aboutoutcome.}
The rule requires a finite verb.
It accepts patterns such as ``we introduce/propose/present'', ``our results show'', and inflected forms of ``achieve'', ``outperform'', and ``improve''.
It rejects navigational patterns such as ``Table/Figure N shows'', future-work cues, method verbs such as ``we use/train/evaluate'' without an outcome cue, and prior-work cues without a current-study reference.

\paragraph{\containsembedded.}
The rule accepts a finite clausal complement (\texttt{ccomp}), a \textit{that}-marked finite clause attached to a content noun such as ``finding'' or ``evidence'', or a post-colon span containing both a subject and a finite verb.

\paragraph{\isreporting.}
The rule requires a matrix verb with a reporting lemma such as ``argue'', ``find'', ``indicate'', ``report'', ``show'', or ``suggest''.
Matches in relative, adverbial, or open complements (\texttt{relcl}, \texttt{acl}, \texttt{advcl}, or \texttt{xcomp}) are excluded; ``find'' must be a root or coordinated root.

\paragraph{\gettargetsent.}
This operation runs only when both \containsembedded and \isreporting are True.
It takes the dependency subtree of the first finite complement, removes an initial ``that'', ``whether'', or ``if'', trims punctuation, and returns the resulting proposition.
Otherwise, it copies the original sentence.

\paragraph{\isgen.}
The rule first accepts causal markers such as ``because'', ``due to'', ``cause'', ``lead to'', and ``result in''.
It then maps vague frequencies such as ``often'', ``usually'', and ``most of the time'' to True, but exact or absolute frequencies such as ``3 out of 10'', ``62\% of cases'', ``always'', and ``never'' to False.
When there is no frequency marker, it approximates \frameworkname's insertion test: a present-tense predication is accepted unless blocked by a study-specific expression (e.g., ``this study'', ``Table 2'', or an exact percentage), a singular proper-name subject, or a past-tense event.

\paragraph{\refstudy.}
The rule accepts explicit current-study expressions: first-person forms such as ``we'' and ``our''; ``this study'' and ``the present study''; ``these results'' and ``these findings''; ``the proposed method''; references such as ``Table/Figure/Section N''; and numbered studies or experiments.

\paragraph{\refstudypop.}
The rule first matches explicit populations such as ``participants'', ``annotators'', ``sample/cohort of'', ``dataset'', ``benchmark'', and ``corpus''.
It also matches experiment--population relations such as ``evaluated \ldots{} models/datasets'', ``across \ldots{} tasks/languages'', or ``on task/benchmark'' when the sentence contains a measurement such as accuracy, BLEU, or an improvement.
Anaphoric expressions such as ``these models'' require an experimental cue in the context.

\paragraph{\ispresenttense.}
The rule inspects the dependency root and its coordinated main predicates.
A \texttt{VBP}/\texttt{VBZ} predicate or present-tense auxiliary is accepted.
The modals ``can'' and ``may'' count as present, whereas ``shall'', ``will'', ``could'', and ``would'' do not.

\paragraph{\containsquant.}
The rule matches phrases such as ``a few'', ``the majority of'', ``some/most/all of'', and ``more/fewer/less than N\%'', as well as tokens such as ``each'', ``every'', ``many'', and ``several'' when they modify a noun.
Adjectival uses such as ``most effective'' are excluded because the token is not attached to a noun.

\paragraph{Evaluation.}
We evaluate each module on the same 200 rows used for human verification.
For each Boolean module, the majority verification judgement supplies the reference label: the \frameworkname label is retained when most annotators judged it correct and inverted otherwise.
For \gettargetsent, all audited extractions were majority-approved, so we report normalised exact match against these reference strings.
The final-category comparison executes the complete baseline on the shared 100-row verification set.
Table~\ref{tab:rule-baseline} reports accuracy alongside the corresponding \frameworkname result; rows whose absolute difference is below $0.1$ are bold.

\begin{table}[t]
    \centering
    \footnotesize
    \setlength{\tabcolsep}{3pt}
    \begin{tabular}{@{}lcc@{}}
        \toprule
        \textbf{Module} & \textbf{\frameworkname} & \textbf{Rule baseline} \\
        \midrule
        \aboutoutcome      & 0.990 & 0.630 \\
        \containsembedded  & 0.925 & 0.810 \\
        \isreporting       & 0.985 & 0.860 \\
        \gettargetsent     & 1.000 & 0.460 \\
        \isgen             & 0.990 & 0.630 \\
        \textbf{\refstudy} & \textbf{0.975} & \textbf{0.895} \\
        \refstudypop       & 0.980 & 0.570 \\
        \textbf{\ispresenttense} & \textbf{0.985} & \textbf{0.965} \\
        \containsquant     & 0.985 & 0.845 \\
        \midrule
        Generalisation category & 0.960 & 0.120 \\
        \bottomrule
    \end{tabular}
    \caption{Verification-derived accuracy of \frameworkname and its prompt-derived rule-based counterparts. Component rows use 200 verified examples; \gettargetsent uses normalised exact match, and the final row uses the shared 100-example verification set.}
    \label{tab:rule-baseline}
\end{table}

The rule baseline approaches \frameworkname on \refstudy and \ispresenttense.
These components could therefore be candidates for deterministic replacement, which would reduce LLM calls and improve reproducibility.
However, the final-category accuracy of $0.120$ shows that wholesale replacement is not viable: semantic distinctions such as generality and references to the study population remain difficult to capture with surface rules, and component errors propagate through the pipeline.
This result aligns with the observation that certain semantic distinctions, particularly genericity, are not transparently marked in English~\citep{reiter-frank-2010-identifying}.
Independent annotation likewise yields very low chance-corrected agreement for \isgen ($\alpha=0.065$) and only moderate agreement for \refstudypop ($\alpha=0.428$; Appendix~\ref{app:independent-annotation}).
This difficulty in reproducing the semantic distinctions partly contextualises the rule baseline's performance.
A hybrid framework combining deterministic rules for suitable components with LLM-based semantic judgements is therefore a promising direction.

We also considered a supervised baseline trained on the 100 non-shared verification rows for each classification step and evaluated on the shared set.
This would not yield a meaningful comparison: beyond the small training set, labels were balanced only across the complete 200-row audit rather than within the non-shared subset.
The resulting training data are severely imbalanced for several steps (e.g., $99{:}1$ for outcome detection and $89{:}11$ for present-tense detection), while the test split is often skewed in the opposite direction.
A learned model's performance would therefore largely reflect the sampling procedure.
The current independent round is also too small to provide an adequately scaled training set.
We therefore leave supervised baselines to future work using a substantially larger, independently labelled dataset.

\section{Dataset Examples and Error Analysis for Annotation}
\label{app:examples-error-analysis}
\definecolor{falseRed}{HTML}{E6308A}
\newcommand{\falsecolour}[1]{\textcolor{falseRed}{#1}}

\definecolor{trueGreen}{HTML}{5BA300}
\newcommand{\truecolour}[1]{\textcolor{trueGreen}{#1}}

\begin{table*}[p]
  \centering
  \footnotesize
  \setlength{\tabcolsep}{3pt}
  \renewcommand{\arraystretch}{1.0}
  \scalebox{0.95}{
  \begin{tabular}{@{}>{\raggedright\arraybackslash}p{0.25\textwidth}>{\raggedright\arraybackslash}p{0.12\textwidth}>{\raggedright\arraybackslash}p{0.45\textwidth}>{\raggedright\arraybackslash}p{0.20\textwidth}@{}}
    \toprule
    Decision path & LLM Label                                                                                                                                                                                                                             & Example ($\top$ = \truecolour{correct}, $\bot$ = \falsecolour{incorrect}) & Error Analysis \\
    \midrule

    \isgen \textit{False}
                  & \ngen
                  & {$\top$ Our method outperforms Set by a maximum of 5.89\% micro F1 in the ZH dataset, highlighting its effectiveness in mitigating order bias.
        \par\noindent\rule{\linewidth}{0.4pt}\par
        $\bot$ This paper detects and classifies fallacies.}
                  & {The $\bot$ example should be annotated as \genast as it should pass the insertion test.}                                                                                                                                                                                                                                          \\
    \hline

    \isgen \textit{True},\newline \refstudypop \textit{True}
                  & \genfng
                  & { $\top$ 73\% of \truecolour{respondents} agree that model-internal interventions are likely necessary to achieve CB.
        \par\noindent\rule{\linewidth}{0.4pt}\par
        $\bot$ There are two advantages brought by such a \falsecolour{data augmentation approach}.}
                  & {The ``data augmentation approach'' in the $\bot$ example is the research outcome rather than the study population.}                                                                                                                                                                                                               \\
    \hline

    \isgen \textit{True},\newline \refstudypop \textit{False},\newline \refstudy \textit{True},\newline \ispresenttense \textit{True},\newline \isreporting \textit{True},\newline \containsembedded \textit{True},\newline \containsquant \textit{True}
                  & \genfng
                  & {$\top$ Our experiments showed that our method offers \truecolour{several} advantages over the prevailing paradigm of direct image-to-report modeling.
        \par\noindent\rule{\linewidth}{0.4pt}\par
        $\bot$ A thorough analysis of the prototype-based clustering method demonstrates that the learned prototype vectors are able to implicitly capture various relations between events.}
                  & {There is no quantifier in the $\bot$ example.}                                                                                                                                                                                                                                                                                    \\
    \hline

    \isgen \textit{True},\newline \refstudypop \textit{False},\newline \refstudy \textit{True},\newline \ispresenttense \textit{True},\newline \isreporting \textit{True},\newline \containsembedded \textit{False}
                  & \genfng
                  & {$\top$ The \truecolour{results indicate} the importance of not freezing too many layers.
        \par\noindent\rule{\linewidth}{0.4pt}\par
        $\bot$ Our extensive experimental results \falsecolour{have demonstrated} the effectiveness of TPT.	}
                  & {The main verb in the $\bot$ example is in the past tense.}                                                                                                                                                                                                                                                                        \\
    \hline

    \isgen \textit{True},\newline \refstudypop \textit{False},\newline \refstudy \textit{True},\newline \ispresenttense \textit{True},\newline \isreporting \textit{False}
                  & \genfng
                  & {$\top$ \truecolour{Our} game \truecolour{requires} more sophisticated analyses to capture to what extent the emergent languages are compositional, and what the decomposed features are.
        \par\noindent\rule{\linewidth}{0.4pt}\par
        $\bot$ In conclusion, offering primary education has few advantages.}
                  & {The $\bot$ example does not refer to the study.}                                                                                                                                                                                                                                                                                  \\
    \hline

    \isgen \textit{True},\newline \refstudypop \textit{False},\newline \refstudy \textit{True},\newline \ispresenttense \textit{False}
                  & \genfng
                  & {$\top$ \truecolour{We} believe that our entropy-based perspective \truecolour{will help} provide a strong starting point for this pursuit.
        \par\noindent\rule{\linewidth}{0.4pt}\par
        $\bot$ By integrating these methods, we achieved a \falsecolour{7.35\% improvement} in decompilation performance, reaching a new state-of-the-art level of \falsecolour{55.03\%}.}
                  & {The $\bot$ should be annotated as \ngen as it should not pass the insertion test. Note that \ispresenttense asserts the main verb of the embedded sentence in the $\bot$ example.}                                                                                                                                                \\
    \hline

    \isgen \textit{True},\newline \refstudypop \textit{False},\newline \refstudy \textit{True},\newline \ispresenttense \textit{True},\newline \isreporting \textit{True},\newline \containsembedded \textit{True},\newline \containsquant \textit{False}
                  & \genfg
                  & {$\top$ Thus, \truecolour{we argue that} annotated arguments \truecolour{are} actually used for AE, not for ED in existing joint methods.
        \par\noindent\rule{\linewidth}{0.4pt}\par
        $\bot$ Note, however, that we should be aware of the possibility of unreliable results from the self-ensemble.}
                  & {The $\bot$ example, although satisfying the designed criteria, its framing is not removable. It shows that our heuristics are not perfect.}                                                                                                                                                                                       \\
    \hline

    \isgen \textit{True},\newline \refstudypop \textit{False},\newline \refstudy \textit{False},\newline \ispresenttense \textit{True},\newline \containsquant \textit{False}
                  & \geng
                  & {$\top$ The contextualized semantic space \truecolour{does not} require additional knowledge, and \truecolour{therefore} can scale to other languages, especially low-resource languages where a large knowledge base is unavailable.
        \par\noindent\rule{\linewidth}{0.4pt}\par
        $\bot$ Contrary to our initial assumption \falsecolour{we} found that the lower the correlation, the higher the predictive performance with FRESH.}
                  & {The $\bot$ exmaple clearly does refer to the study.}                                                                                                                                                                                                                                                                              \\
    \hline

    \isgen \textit{True},\newline \refstudypop \textit{False},\newline \refstudy \textit{False},\newline \ispresenttense \textit{True},\newline \containsquant \textit{True}
                  & \geno
                  & {$\top$ \truecolour{Many} novel methods \truecolour{are} proposed based on IA findings and highly influenced by them, but highly influential non-IA work cites IA findings without being driven by them.
        \par\noindent\rule{\linewidth}{0.4pt}\par
        $\bot$ Such a workflow can provide a \falsecolour{large} improvement in translation quality at each iteration}
                  & {``large'' in the $\bot$ example is not a frequency expression. }                                                                                                                                                                                                                           \\
    \hline

    \isgen \textit{True},\newline \refstudypop \textit{False},\newline \refstudy \textit{False},\newline \ispresenttense \textit{False}
                  & \geno
                  & {$\top$ In addition, L2 regularization \truecolour{worked} better than L0 .
        \par\noindent\rule{\linewidth}{0.4pt}\par
        $\bot$ Effectiveness of Proof-Enhancement Method.}
                  & {Incomplete sentences such as the $\bot$ example should have been discarded during \aboutoutcome stage.}                                                                                                                                                                                                                                            \\
    \bottomrule
  \end{tabular}
  }
  \caption{Example and error analysis for each possible decision path. The $\top$ examples illustrate typical sentences that are expected to fall into the corresponding decision path, while the $\bot$ examples illustrate potential LLM errors and limitations of the current heuristics.}
  \label{tab:path-error-analysis}
\end{table*}

Table~\ref{tab:path-error-analysis} gives representative examples for every terminal path in the annotation decision tree.
For each path, the $\top$ example shows a sentence whose classifier decisions support the assigned taxonomy label, while the $\bot$ example illustrates either a plausible LLM error or a limitation of our heuristic operationalisation.

Several error patterns recur across paths.
First, some mistakes arise at the initial generalisation boundary: result claims can be incorrectly treated as \ngen, while highly specific numerical results may be over-classified as \genast.
Second, the boundary between study framing and study population references is delicate.
For example, phrases such as ``data augmentation approach'' may look like a study population but actually name the research outcome itself.
Third, genericity decisions are sensitive to the exact predicate being evaluated after a reporting frame is removed.
Errors occur when the model misses tense, treats vague descriptors (e.g., ``large'') as quantifiers, or evaluates the surface reporting clause rather than the embedded proposition.

The table also highlights a residual limitation of the decision-tree approximation.
Some sentences satisfy the observable criteria used by the prompts but remain difficult to classify under the theoretical taxonomy, especially when the study frame is pragmatically non-removable even though it has the right syntactic shape.
We therefore use these examples as qualitative diagnostics for the pipeline, complementing the human-validation results in Appendix~\ref{app:validation_details}.

\section{\aboutoutcome Across Sections}
\label{appendix:about_outcome_per_section}

In Table \ref{tab:about_outcome_per_section} we report how many sentences across each section ended up with an \aboutoutcome annotation (note that this is not an average per paper, rather, we normalise by the total set of all sentences from all papers per venue) in Table \ref{tab:about_outcome_per_section}.
Most of the sections are relatively consistent across the tested venues, with the biggest range in the rates of generalisations across the conclusion.

\begin{table}[!htb]
\small
\begin{tabular}{cccccc} \toprule
 &
  \begin{tabular}[c]{@{}c@{}}\textsc{acl}\\ {\small 2017}\end{tabular} &
  \begin{tabular}[c]{@{}c@{}}\textsc{coling}\\ {\small 2020}\end{tabular} &
  \begin{tabular}[c]{@{}c@{}}\textsc{arr}\\ {\small 2022}\end{tabular} &
  \begin{tabular}[c]{@{}c@{}}\textsc{emnlp}\\ {\small 2023}\end{tabular} &
  \begin{tabular}[c]{@{}c@{}}\textsc{emnlp}\\ {\small 2024}\end{tabular} \\ \midrule
\textit{abstract}     & 57.3 & 48.3 & 54.9 & 55.1 & 53.5 \\
\textit{intro}        & 32.2 & 29.5 & 35.1 & 37.8 & 37.4 \\
\textit{results}       & 55.5 & 60.7 & 60.1 & 57.2 & 62.8 \\
\textit{discuss.}     & 56.1 & 64.1 & 54.0 & 54.2 & 57.6 \\
\textit{concl.}       & 55.8 & 55.0 & 36.3 & 71.5 & 34.1 \\ \midrule
\textit{total}       & 51.4 & 51.5 & 48.1 & 55.2 & 49.1 \\ 
\bottomrule
\end{tabular}
\caption{Percentage of \aboutoutcome across different paper sections (across total set of all sentences in each venue from a specific section).}
\label{tab:about_outcome_per_section}
\end{table}

\section{Track Assignment}
\label{appendix:tracks}

The only venue in our dataset which came with track data for each paper was COLING-2020, with the tracks summarised in Table \ref{tab:coling_tracks}.

\begin{table}[!htb]
\centering
\footnotesize
\begin{tabular}{ll}
\toprule
\textbf{Short Name} & \textbf{Full Track} \\
\midrule
Applications     & Applications (BioNLP/Legal/Social...) \\
Dialogue         & Dialogue and Interactive Systems \\
Discourse        & Discourse and Pragmatics \\
TopicModel       & Topic Modelling, IR \& CR \\
LangGen          & Language Generation \\
LangModelling    & Language Modelling \\
KG               & Info Extraction, ..., KGs \\
LR-Eval          & Language Resources and Eval \\
CogMod           & Ling Theories, Cognitive Models \\
ML               & Machine Learning for CL/NLP \\
MT               & Machine Translation \\
WordSeg          & Morphology and Word Segmentation \\
Semantics        & Semantics (Words/Sents/Ontologies) \\
SA-Opinion       & Sentiment Analysis, Opinion Mining \\
Speech           & ASR, TTS \& Spoken Language \\
Summar-Simpl     & Summarization and Simplification \\
SyntaxParsing    & Tagging/Chunking/Syntax Parsing \\
Text-QA          & Textual Inference and QA \\
GroundLA         & Grounded Language Acquisition \\
\bottomrule
\end{tabular}
\caption{COLING 2020 tracks, both full name and our shortened acronyms for plots. For more track information see the call for submissions: \url{https://groups.google.com/g/ml-news/c/pP0wKlJdW6s?pli=1}}
\label{tab:coling_tracks}
\end{table}

For all COLING-2020 papers, we got an embedding of the title + abstract using the \verb|allenai/specter2_base| pre-trained model \citep{specter2}.
We then obtained centroid vectors for all tracks by considering all COLING-2020 embeddings in each one.
We then attached a soft track label to all track-less papers from the other four venues by embedding their title + abstract using the same \verb|allenai/specter2_base| pre-trained model and simply labelling the paper as belonging to the closest track (lin-alg norm for distance from track centroids).

\begin{figure*}[!ht]
\centering
\includegraphics[width=0.90\textwidth]{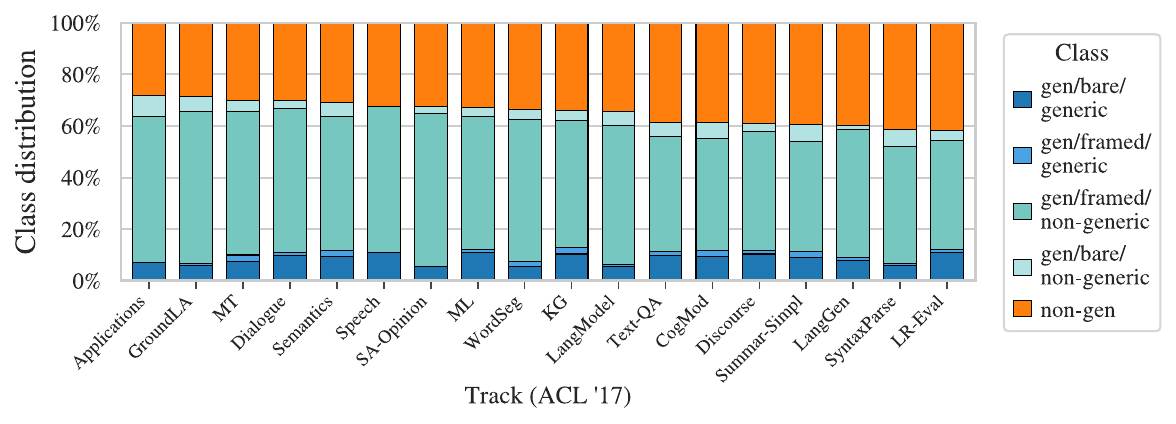}
\caption{Generalisation rates across ACL '17 papers with soft track labels.}
\label{fig:gen_class_tracks_ACL_17}
\end{figure*}

\begin{figure*}[!ht]
\centering
\includegraphics[width=0.90\textwidth]{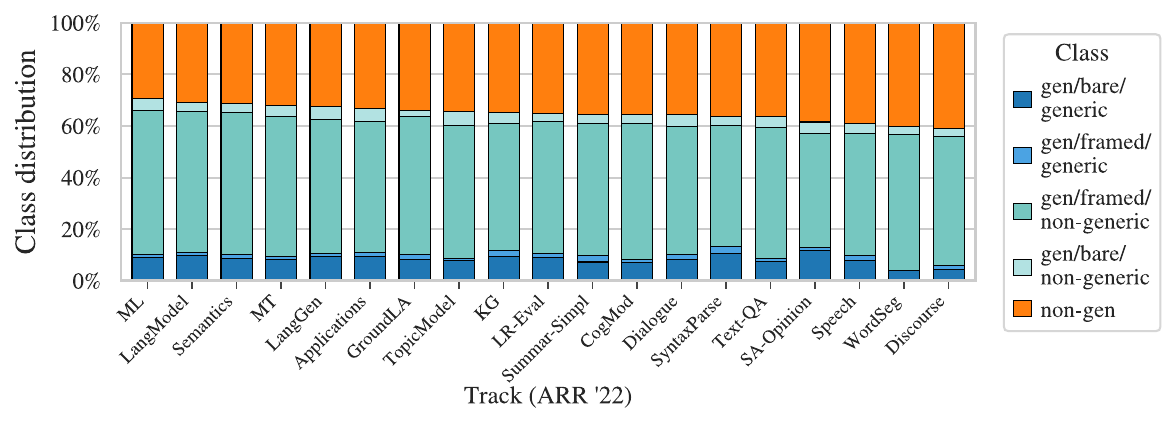}
\caption{Generalisation rates across ARR '22 papers with soft track labels.}
\label{fig:gen_class_tracks_ARR_22}
\end{figure*}

\begin{figure*}[!ht]
\centering
\includegraphics[width=0.90\textwidth]{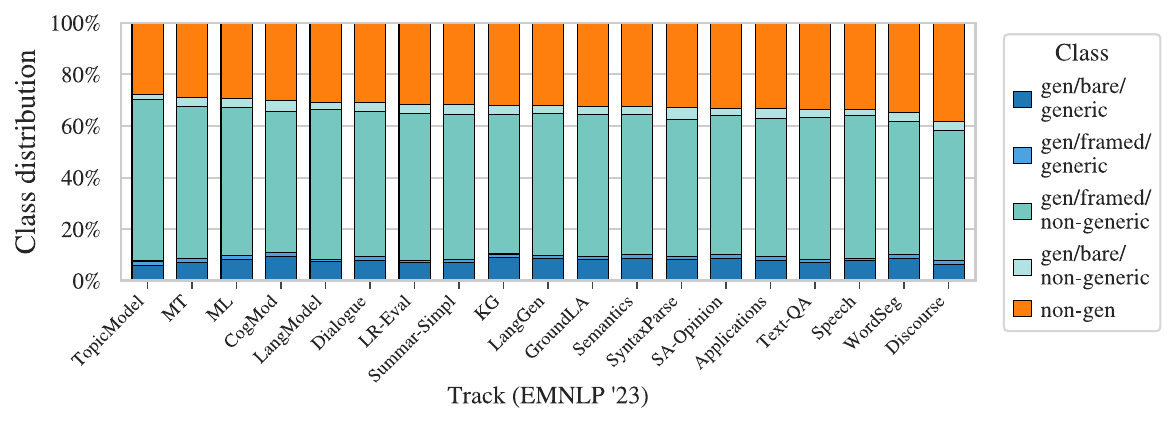}
\caption{Generalisation rates across EMNLP '23 papers with soft track labels.}
\label{fig:gen_class_tracks_EMNLP_23}
\end{figure*}

\begin{figure*}[!ht]
\centering
\includegraphics[width=0.90\textwidth]{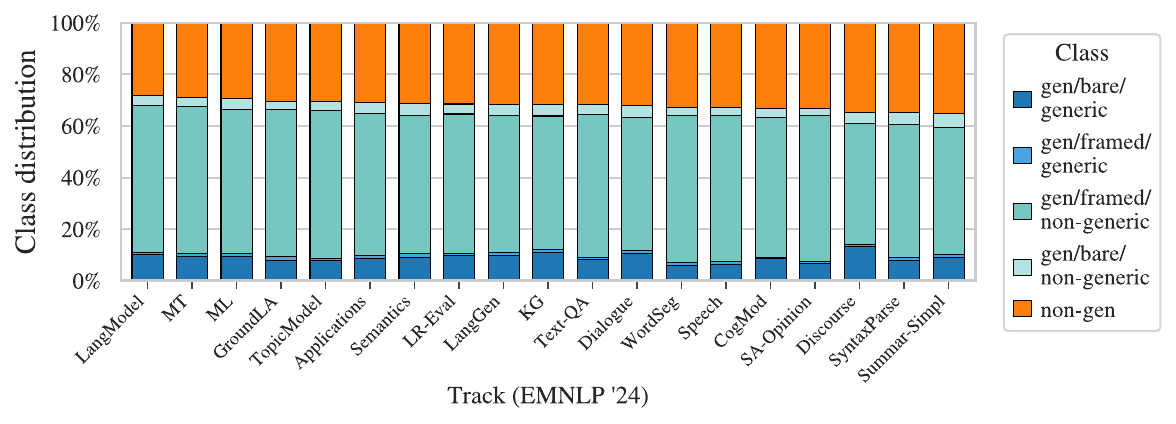}
\caption{Generalisation rates across EMNLP '24 papers with soft track labels.}
\label{fig:gen_class_tracks_EMNLP_24}
\end{figure*}

We then plot the generalisation class distirbution across the different venues, separating results across papers assigned to the same track.
ACL '17 in Figure \ref{fig:gen_class_tracks_ACL_17},
ARR '22 in Figure \ref{fig:gen_class_tracks_ARR_22},
EMNLP '23 in Figure \ref{fig:gen_class_tracks_EMNLP_23},
EMNLP '24 in Figure \ref{fig:gen_class_tracks_EMNLP_24}.
We find that rates of generalisations do differ across the different NLP subdomains, although the trends are not consistent across the venues.
Additionally, the relative orderings of the tracks in terms of most to least generalised are not consistent across the different years.
There are several explanations to this.
First, the NLP field changed quite substantially over the past couple of years, so some of the original COLING-2020 tracks are not as applicable anymore.
Additionally, our soft-labelling clustering algorithm is quite naive, and although we manually inspected the soft labels for each venue to confirm that something sensible was going on, the approach is far from perfect.
We believe it is still useful to consider different rates of generalisations across different scientific subfields, but perhaps more gold labels for tracks/domains are necessary for a robust analysis.

\section{Citation Analysis}
\label{appendix:citations}

To obtain citation counts, we first parse each paper's ID from the ACL Anthology URL (if provided) and then pass this to the SemanticScholarAPI\footnote{\url{https://api.semanticscholar.org/graph/v1/paper/batch}}.
Note that we were not able to obtain citation counts for papers that did not contain the ACL URL in their metadata, which was the case for EMNLP '23 papers.
In Table \ref{tab:spearman_rho_per_venue} we report the results of the same analysis as was done for ACL '17 papers (obtaining the Spearman rho ($\rho$) correlation coefficient between generalisation classes and paper citation counts) for COLING '20, ARR '22, and EMNLP '24 papers.

\begin{table*}[!htb]
\centering
\small
\begin{tabular}{lcccccc}
\toprule
 & \multicolumn{2}{c}{COLING '20} & \multicolumn{2}{c}{ARR '22} & \multicolumn{2}{c}{EMNLP '24} \\
 & \multicolumn{1}{c}{$\rho$} & \multicolumn{1}{c}{p-value} & \multicolumn{1}{c}{$\rho$} & \multicolumn{1}{c}{p-value} & \multicolumn{1}{c}{$\rho$} & \multicolumn{1}{c}{p-value} \\ \midrule
\ngen & -0.07 & 0.541 & -0.06 & 0.168 & -0.02 & 0.365 \\
\genfng & 0.14 & 0.218 & 0.08 & 0.070 & -0.01 & 0.623 \\
\geng & -0.04 & 0.712 & 0.03 & 0.538 & 0.02 & 0.321 \\
\geno & -0.06 & 0.587 & -0.07 & 0.107 & 0.01 & 0.584 \\
\genfg & 0.04 & 0.741 & -0.05 & 0.287 & 0.04 & 0.111 \\ \midrule
\generic & -0.03 & 0.769 & 0.00 & 0.933 & 0.02 & 0.367 \\
\genast & 0.07 & 0.541 & 0.06 & 0.168 & 0.02 & 0.361 \\
\genfast & 0.14 & 0.199 & 0.08 & 0.088 & -0.01 & 0.648 \\ \bottomrule
\end{tabular}
\caption{Spearman $\rho$ correlation between the generalisation classes and citation counts for COLING '20, ARR-22, EMNLP '24 papers.}
\label{tab:spearman_rho_per_venue}
\end{table*}

\begin{figure*}[!ht]
\centering
\includegraphics[width=0.85\textwidth]{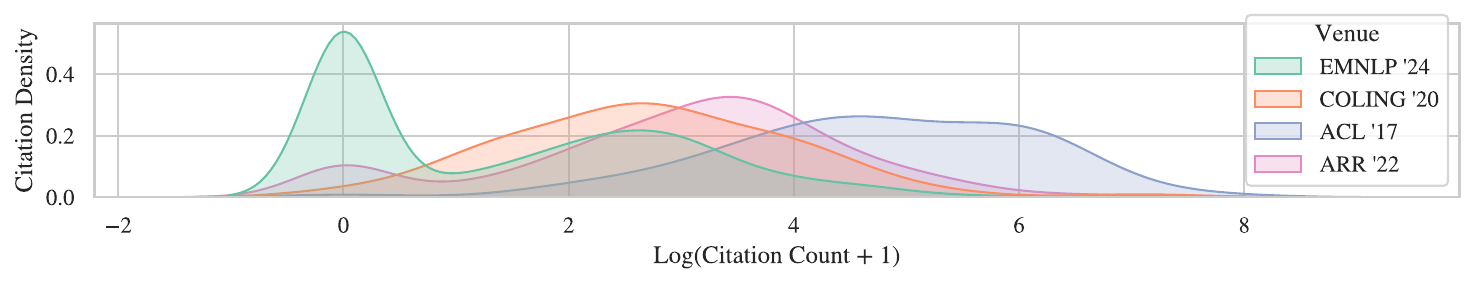}
\caption{Kernel Density Estimation (KDE) plot of the citation distributions for papers across the venues for which we could obtain citation counts using the Semantic Scholar API (ACL '17, COLING '20, ARR '22, EMNLP '24).}
\label{fig:citation_densities}
\end{figure*}

Unlike for ACL '17, none of the correlations are statistically significant.
However, we still see the \genfast class as one of the more positively correlated ones across COLING '20 and ARR '22, which is consistent with what we observed with ACL '17, even though the effects are not strong enough statistically.
As noted in the main text, correlating citation counts with a single linguistic feature, without any consideration for the semantics of the paper's claims, is an incredibly hard task.
Therefore, the fact that we are seeing even a weak signal is already an indication that this relationship should be robustly investigated further, perhaps alongside some combination of semantic features.
Also, the different venues have really different citation distributions (see Figure \ref{fig:citation_densities}).
ACL '17 specifically has more spread, whereas for the other venues either the citation counts have not stabilised (for EMNLP '24, see the sharp spike around 0 citations), or the citation counts are all all bundled in the same log space.
It is possible that the distributions of the other conferences simply are not spread out enough and are too clustered/noisy to find a strong effect.

\section{Data Access, Licensing, and Intended Use}

We build \datasetname on the NLPeer corpus~\citep{dycke-etal-2023-nlpeer}, accessed by submitting the standard data request through the official TUdatalib deposits for v1\footnote{\url{https://tudatalib.ulb.tu-darmstadt.de/items/eeb96857-15ea-4075-9857-2f98a9728145}} and v2\footnote{\url{https://tudatalib.ulb.tu-darmstadt.de/items/d4a4061b-e4e3-4b1e-a90d-d48a3d69e3c0}}; no separate data use agreement was attached.
Both deposits are released under \href{https://creativecommons.org/licenses/by-nc/4.0/}{CC BY-NC 4.0} except where otherwise noted, with subset-level Creative Commons terms (e.g., CC-BY for ACL-17, CC-BY-NC-SA 4.0 for COLING-20 and ARR-22) and per-item licence strings documented in the resource paper~\citep{dycke-etal-2023-nlpeer} and shipped as metadata with each paper.
Our use is non-commercial academic research with attribution, and we do not redistribute the raw corpus.
We modify the source material only by parsing manuscripts into intertextual graphs, segmenting them into sentences, filtering to five standard sections, and adding sentence-level annotations.

NLPeer is intended for the computational study of peer review and NLP-based reviewing assistance, and the authors stress that author or reviewer profiling violates the intended use; they explicitly position NLPeer as a basis for new annotated datasets on its clearly licensed, structured manuscript text~\citep{dycke-etal-2023-nlpeer}.
Our work falls within this scope: we process only parsed manuscript text, annotating linguistic properties of scientific claims. We do not use peer-review text, attempt profiling, or automate review generation or scoring.
We will release the resulting \datasetname annotations under \href{https://creativecommons.org/licenses/by-nc-sa/4.0/}{CC BY-NC-SA 4.0} to inherit the most restrictive source-subset licence, with NLPeer attributed and the per-row source licence retained alongside each annotation.

\section{Personal Data and Offensive Content}
We use manuscript text only and do not access peer-review reports, reviewer identities, or any non-public author metadata.
Identifying information about reviewers in NLPeer was already curated upstream: review texts were manually checked by the NLPeer team for personal information, and identifying metadata was stripped from review reports in all subsets except the openly identified F1000-22~\citep{dycke-etal-2023-nlpeer}; F1000-22 is excluded from our work.
Author names that appear in the manuscripts (e.g., in ``we''-statements, acknowledgements, or in-text citations) reflect public scholarly publication records rather than private identifiers we collected; we never use them for profiling, demographic inference, or any analysis at the level of individual people.
Beyond venue publication norms, we did not perform a separate systematic screen for offensive content; given that the input is accepted NLP/CL research papers, we judged this risk to be low.

\section{Prompt Templates}
\onecolumn

\begingroup\hypersetup{linkcolor=white}
\begin{prompt}[title={\aboutoutcome\hfill{}(\S\ref{subsec:isoutcome})}]
\# Task: \\
Determine if the target claim is an outcome, given the previous two sentences as context. When you are not sure, answer with Yes. \\
 \\
\# Outcome: \\
* Introduction, e.g. \\
  * "we introduce", "we propose", "we present" \\
* Results: observations of the experimental results, e.g. \\
  * "Our model achieved significant improvements over the baselines." \\
* Discussion: interpretation or analysis of the results, e.g. \\
  * "The results show that our approach significantly outperforms baseline methods." \\
* Contribution: how the research will contribute to knowledge in the field, e.g. \\
  * "This work contributes to the field by introducing a novel architecture." \\
* Conclusion \\
* Limitations \\
 \\
\# Not Outcome: \\
* Background: statements about knowledge in the domain or previous related work, e.g. \\
  * "Deep learning models achieve high accuracy in most tasks." \\
* Experimental setup details, e.g. \\
  * "We used a BM25 retriever with a top $k$ of $4$ and a width penalty $\beta=0.6$." \\
* Methodology/Approach, including actions done by the authors, e.g. \\
  * "We evaluate the performance of our method on various benchmarks." \\
  * "GEAR takes both prediction and evaluation metrics as input." \\
  * "We believe that GEAR is able to take both prediction and evaluation metrics as input." \\
* Navigational sentences: referring to text, figures, tables, etc., e.g. \\
  * "Table 1 shows the results of the experiments." \\
* Incomplete claims or parsing errors, e.g. \\
  * "in both the pre-training and the fine-tuning stages." \\
  * "2 3 5 5 1 ] ] ] " \\
* Future work \\
 \\
\# More instructions: \\
* Outcome should be mainly about the author's own work. It should not be fully about others' work. \\
* It can be an outcome if others' work is only mentioned to elaborate on the findings of the authors, e.g. \\
  * "Unlike previous works, GEAR nails single-stage summarisation, aligning with the finding of Gu et al. (2024)." \\
* Modal verbs or hedging (e.g., "can", "may", "tend to") do not change the classification. Classify by the content. \\
* Being an opinion (e.g., "we believe") does not change the classification. \\
* If both outcome and non-outcome are present, choose outcome, e.g. \\
  * "We propose the method to address the limitations by introducing a novel architecture." \\
 \\
\# Target Claim: \\
Context: \{two\_sent\_context\} \\
Claim: \{claim\} \\
 \\
Is the target claim an outcome, given the previous two sentences as context? When you are not sure, answer with Yes. Give your reasoning first and then answer with Yes or No. The output should be in the format below. \\
Reasoning: <your reasoning> \\
Answer: Yes/No \\
\end{prompt}

\begin{prompt}[title={\isgen\hfill{}(\S\ref{subsec:isgen})}]
\# Task: Determine if the target claim is generalised. Answer with generalised or not generalised. \\
 \\
\# Decision process: \\
1. Check the context \\
The context provided is the previous two sentences before the target claim. \\
 \\
2. Ignore hedging \\
Do not let hedging affect the decision. \\
Examples of hedging: can, could, would, may, might, appears to, seem, likely, necessarily. \\
 \\
3. Check for explicit causality \\
Check whether the sentence explicitly expresses causality, for example by using words or phrases such as: because, reason, due to, so, lead to, result in, cause. \\
If YES -> generalised \\
If NO -> continue \\
Example: \\
"The improvement is due to better attention alignment." -> generalised \\
 \\
4. Check for a frequency expression in the whole sentence \\
4.1. If there is no frequency expression \\
-> continue to 5 \\
 \\
4.2. If there is already a frequency expression, check whether it expresses or implies one of the following: \\
* a specific frequency, e.g. "3 out of 10", "in 62\% of cases" \\
* 100\% frequency, e.g. "in all cases", "always", "all the time", "consistently" \\
* 0\% frequency, e.g. "never", "no", "none" \\
If YES -> not generalised \\
If NO -> generalised \\
Examples: \\
"The model succeeds half of the time." \\
-> not generalised \\
"The model always outperforms the baseline." \\
-> not generalised \\
"The model often/usually/sometimes outperforms the baseline." \\
-> generalised \\
"The model outperforms the baseline most of the time/more than half of the time." \\
-> generalised \\
 \\
5. If there is no frequency expression, apply the insertion test \\
Try inserting either "usually" or "generally" into the **main clause**, taking the context into account. \\
Then ask: Does the modified sentence still sound natural in the context? \\
If the modified sentence still sounds natural in the context -> generalised \\
If the modified sentence does not sound natural in the context -> not generalised \\
 \\
Notes on interpretation \\
* "generally" ONLY allows a collective reading \\
  * the property applies to the subject as a class/kind or whole, not to a single instance \\
* "usually" can allow: \\
  * a distributive reading: indicates the proportion of subject instances that the property applies to, e.g. \\
  "The scores of A are usually lower than B" \\
  * a temporal reading: indicates the proportion of time that the property applies, e.g. \\
  "A usually works better than B" \\
* Use whichever reading fits the context best. \\
 \\
6. Mixed cases \\
If a sentence contains both a generalised part and a non-generalised part, annotate it as generalised. \\
Example: \\
"On this dataset, the model consistently achieved 85\% accuracy; this shows that graph structure helps." \\
-> contains both 100\% frequency ("consistently") and a generalised claim ("graph structure helps") \\
-> generalised \\
 \\
\# Decision summary \\
Use this decision order: \\
* Read the previous three sentences for context. \\
1. Read the previous three sentences for context. \\
2. Ignore hedging. \\
3. If it explicitly expresses causality -> **generalised**. \\
4. Otherwise, check for a frequency expression in the whole sentence: \\
  * no frequency expression -> continue to 5 \\
  * specific / 100\% / 0\% frequency -> **not generalised** \\
  * other frequency expressions -> **generalised** \\
5. If there is no frequency expression, try inserting "usually/generally" into the main clause: \\
  * the modified sentence still sounds natural in the context -> **generalised** \\
  * the modified sentence does not sound natural in the context -> **not generalised** \\
6. If both types are present, choose **generalised**. \\
 \\
\# Target Claim: \\
context: \{two\_sent\_context\} \\
claim: \{get\_target\_sent\_label\} \\
Following the instructions above, is the target claim generalised? Give the reasoning process first and then answer with "generalised" or "not generalised". The output should be in the format below. \\
Reasoning Process: \\
  * explicit causality exists: Yes/No \\
  * given no explicit causality, frequency expression exists: Yes/No/NA \\
  * given frequency expression exists, it is a specific / 100\% / 0\% frequency: Yes/No/NA \\
  * given frequency expression exists, it is other frequency expression: Yes/No/NA \\
  * given no frequency expression, adding "usually/generally" into the main clause sounds natural: Yes/No/NA \\
Answer: generalised/not generalised \\
\end{prompt}

\begin{prompt}[title={\refstudy\hfill{}(\S\ref{subsec:isframe})}]
\# Task: Determine if the target claim includes a reference to the study itself. Find the reference first, if there is any, and then answer with Yes or No, in the same format as the examples. \\
 \\
\# Instruction: \\
The task is to determine whether the target claim is explicitly portrayed as a fact that the current study conducted by the authors tells us. Such references could include, but are not limited to: \\
* the authors, e.g. "we", "our" \\
* the research findings or outcomes, e.g. "the finding", "the proposed method" \\
* experimental results, e.g. "the experiment results", "the table shows" \\
* direct reference to the study, e.g. "study 2", "(the third experiment)" \\
* elements of the paper, e.g. "Table 1", "Figure 5", "Section 3", "top 5 rows of Table 1" \\
 \\
\# Target Claim: \\
Claim: \{get\_target\_sent\_label\} \\
 \\
Determine if the target claim includes a reference to the study itself. Find the reference first (no more than 3 words), if there is any, and then answer with Yes or No, in the format as below. \\
Reference: <reference to the study from the target claim> \\
Answer: Yes/No \\
\end{prompt}

\begin{prompt}[title={\refstudypop\hfill{}(\S\ref{subsec:isframe})}]
\# Task: \\
Determine whether the target claim, based on the context, identifies the population that the experiments are carried out on. If it does, state the population. \\
 \\
\# Instructions: \\
Consider the question: "What are the experiments carried out on?" A valid answer may include, but is not limited to: \\
* datasets, e.g. "the MuSE dataset", "SemEval-2016 Task 10", "the top 5 rows" \\
* models, e.g. "the language models we evaluate", "all auto-regressive models we tested", "we testedGPT-4-128k" \\
* participants, e.g. "participants", "annotators", "markers", "students" \\
 \\
\# Target Claim: \\
Context: \{two\_sent\_context\} \\
Claim: "\{get\_target\_sent\_label\}" \\
 \\
Determine whether the target claim, based on the context, could answer the question "What are the experiments carried out on?" State any population you find, or "none", then answer with "Yes" or "No". Use the format below: \\
Population: <population identified in the claim, or "none"> \\
Answer: Yes/No \\
\end{prompt}

\begin{prompt}[title={\ispresenttense\hfill{}(\S\ref{subsec:isgeneric})}]
\# Task: Determine if the main clause of the target claim is in the present tense. State the main verb in the main clause, then answer with "Yes" or "No". \\
 \\
\# Instructions: \\
* Predication combined with present tense and other tenses should be counted as present tense. For example, "it passed the test and shows great improvement" should be classified as present tense. \\
* Treat "can" and "may" as present tense. \\
* Treat "shall" and "will" as future tense. \\
* Treat "could", "would" as past tense \\
 \\
\# Examples: \\
Claim: "it passed the test and shows great improvement" \\
Main verb: "passed", "shows" \\
Answer: Yes \\
 \\
Claim: Adult's performance on a spatial reasoning task was not related to their scores on a verbal reasoning task \\
Main verb: "was" \\
Answer: No \\
 \\
Claim: The three groups did not differ in their ability to discriminate non-modified word \\
Main verb: "did not differ" \\
Answer: No \\
 \\
\# Target Claim: \\
Claim: \{get\_target\_sent\_label\} \\
 \\
Determine whether the target claim is in the present tense. State the main verb in the main clause, then answer with "Yes" or "No". The output should be in the format below. \\
Main verb: <main verb in the main clause> \\
Answer: Yes/No \\
\end{prompt}

\begin{prompt}[title={\containsquant\hfill{}(\S\ref{subsec:isgeneric})}]
\# Task: Determine whether the target claim includes non-numeric quantifiers. \\
 \\
\# Definition of Quantifiers: \\
Quantifiers are words that limit how broadly a claim applies. This includes words that specify the extent, frequency, or proportion of the claim. \\
 \\
\# Examples of Quantifiers: \\
* Universal / distributive: all, both, each, every \\
* Existential / indefinite: some, any, several, a few, few, many, much, little, a little, a lot of, lots of, plenty of \\
* Negative: no, none, neither, hardly any \\
* Proportional: most, half (of), the majority of, the minority of \\
* Comparative / superlative before **nouns**: more (than), most, fewer (than), fewest, less (than), least \\
* Partitives (of-phrases): some of, most of, all of, none of, many of, much of, a lot of, plenty of \\
 \\
\# Not Quantifiers: \\
* Anything that is not before a noun is not a quantifier. \\
* Specific numerals are not quantifiers, e.g. "0.9", "70\%". \\
 \\
\# More Examples: \\
Claim: Most reasons do not adequately justify violating a sacred value. \\
Reasoning: In this case, "most" is before a noun, hence it is a quantifier. \\
Answer: Yes \\
 \\
Claim: The algorithm is the most effective one. \\
Reasoning: "most" here is before an adjective, not a noun, hence it is not a quantifier. \\
Answer: No \\
 \\
Claim: Children generally score high on measures of academic achievement. \\
Reasoning: "generally" is not before a noun, it is not a quantifier. \\
Answer: No \\
 \\
Claim: While single-head attention is 0.9 BLEU worse than the best setting, quality also drops off. \\
Reasoning: Specific numerals are not quantifiers. \\
Answer: No \\
 \\
Claim: Children score highest on measures of academic achievement. \\
Reasoning: "highest" here is not before a noun, hence it is not a quantifier. \\
Answer: No \\
 \\
Claim: The algorithm achieves more than 95\% accuracy. \\
Reasoning: Although specific numerals "95\%" is not a quantifier, "more than" here is before a noun, hence it is a quantifier. \\
Answer: Yes \\
 \\
\# Target Claim: \\
\{get\_target\_sent\_label\} \\
 \\
Determine whether the target claim includes non-numeric quantifiers. Give your reasoning first, then answer with Yes or No. Use the format below. \\
Reasoning: <your reasoning> \\
Answer: Yes/No \\
\end{prompt}

\begin{prompt}[title={\isreporting\hfill{}(\S\ref{subsec:isgeneric})}]
\# Task: Determine whether the target claim contains a reporting verb. \\
 \\
\# Instructions: \\
* A reporting clause is a clause that indicates you are referring to what someone said or thought. For example, in "She said that she was hungry", "She said" is a reporting clause.  \\
* In the context of academic writing, a reporting clause is often used to report the findings of a study. For example, "We found ...", "the results demonstrate ...", and "it suggests ..." are reporting clauses. \\
* A reporting verb is the main verb in the reporting clause, e.g. "found", "demonstrate", "suggest". \\
 \\
\# Examples: \\
Claim: Our results speak to the importance of psychological distancing in the expression of conscious control over thought and action from a young age \\
Reasoning: "speak" is a reporting verb \\
Answer: Yes \\
 \\
Claim: The finding that the developing system favours risky visuomotor choices forms a first step towards understanding how children deal with everyday activities \\
Reasoning: "forms" is not a reporting verb \\
Answer: No \\
 \\
Claim: Children's ability to improve executive function by mentally transcending their egocentric perspective underscores the critical role of representational capacities in self-control \\
Reasoning: "underscores" is a reporting verb \\
Answer: Yes \\
 \\
Claim: Across four studies it shows that people do better on shape-recognition tasks \\
Reasoning: "shows" is a reporting verb \\
Answer: Yes \\
 \\
Claim: the MedBERT model achieves state-of-the-art results for multiple tasks. \\
Reasoning: "achieves" is not a reporting verb \\
Answer: No \\
 \\
\# Target Claim: \\
Claim: \{claim\} \\
 \\
Determine whether the target claim contains a reporting verb. Give your reasoning first, and then answer with Yes or No. The output should be in the format below. \\
Reasoning: <your reasoning> \\
Answer: Yes/No \\
\end{prompt}

\begin{prompt}[title={\containsembedded\hfill{}(\S\ref{subsec:isgeneric})}]
\# Task: Determine whether the target claim contains a complement clause. \\
 \\
\# Instructions: \\
A complement clause can be distinguished from a relative clause because a relative complementiser can be replaced by a relative wh-pronoun (e.g. who or which), whereas a complement clause cannot be replaced in this way. \\
 \\
\# Examples: \\
Claim: The present study confirms that there exist significant associations between musical practice and intelligence. \\
Reasoning: “The present study confirms which there exist significant associations between musical practice and intelligence” does NOT make sense. \\
Answer: Yes \\
 \\
Claim: These findings suggest a promising new avenue for early EF intervention. \\
Reasoning: There is no subordinate clause in the sentence; the part after “suggest” is a noun phrase. \\
Answer: No \\
 \\
Claim: The present study confirms the existence of significant associations between musical practice and intelligence. \\
Reasoning: There is no subordinate clause in the sentence; the part after “confirms” is a noun phrase. \\
Answer: No \\
 \\
Claim: These findings suggest that this is a promising new avenue for early EF intervention. \\
Reasoning: “These findings suggest which this is a promising new avenue for early EF intervention” does NOT make sense. \\
Answer: Yes \\
 \\
Claim: Our results speak to the fact that psychological distancing is important in the expression of conscious control over thought and action from a young age. \\
Reasoning: “Our results speak to the fact which psychological distancing is important in the expression of conscious control over thought and action from a young age” does NOT make sense. \\
Answer: Yes \\
 \\
Claim: Our results speak to the importance of psychological distancing in the expression of conscious control over thought and action from a young age. \\
Reasoning: There is no subordinate clause in the sentence; the part after “speak to” is a noun phrase. \\
Answer: No \\
 \\
Claim: Our results show L1 loss is not a crucial condition for acquiring an L2. \\
Reasoning: “Our results show which L1 loss is not a crucial condition for acquiring an L2” does NOT make sense. \\
Answer: Yes \\
 \\
\# Target Claim: \\
\{claim\} \\
 \\
Does the target claim contain a complement clause? Give your reasoning first, then answer with Yes or No. Use the format below. \\
Reasoning: <your reasoning> \\
Answer: Yes/No \\
\end{prompt}

\begin{prompt}[title={\gettargetsent\hfill{}(\S\ref{subsec:isgeneric})}]
\# Task \\
You will be given a reporting clause (e.g., "this table suggests...", "we found...", "it shows..."). Your task is to analyse it and extract the embedded sentence (complement clause) being reported. If the sentence is not a reporting clause, or if the reported content is not a complete sentence, return the original sentence. \\
 \\
\# Instructions \\
* Possible reporting verbs include, but are not limited to: suggest, indicate, demonstrate, show, argue, highlight, report, note, observe, conclude, state, claim, illustrate, examine, propose, and find. \\
* The reported content should be a complete sentence. \\
 \\
\# Examples \\
Sentence: This table suggests that each handcrafted feature contributes to some extent, whereas features with large weights vary across prompts. \\
Reasoning: "suggests" is a reporting verb, and the reported content is a complete sentence, so we extract it; the "whereas" clause should also be included. \\
Answer: Each handcrafted feature contributes to some extent, whereas features with large weights vary across prompts. \\
 \\
Sentence: Table 4 shows that by incorporating handcrafted essay-level features, the proposed method drastically improves the accuracy of all base DNN-AES models. \\
Reasoning: "shows" is a reporting verb, and the reported content is a complete sentence, so we extract it; the "by incorporating" clause should also be included. \\
Answer: By incorporating handcrafted essay-level features, the proposed method drastically improves the accuracy of all base DNN-AES models. \\
 \\
Sentence: Table 4 shows the experimental results. \\
Reasoning: "shows" is a reporting verb, but the reported content is not a complete sentence, so we return the original sentence. \\
Answer: Table 4 shows the experimental results. \\
 \\
Sentence: This task presents specific challenges: the output is subject to strong structural constraints and is significantly longer than the input. \\
Reasoning: "presents" functions as a reporting verb here, and the reported content is a complete sentence, so we extract it. \\
Answer: The output is subject to strong structural constraints and is significantly longer than the input. \\
 \\
Sentence: Another reason is that reducing the attention key size hurts model quality. \\
Reasoning: "is" is not a reporting verb, so we return the original sentence. \\
Answer: Another reason is that reducing the attention key size hurts model quality. \\
 \\
Sentence: The results, shown in the "p-value" column in Table 4, indicated that the proposed method improved performance at the 5\% significance level for the LSTM- and BERT-based models, and at the 10\% significance level for the conventional hybrid model. \\
Reasoning: "indicated" is a reporting verb, and the reported content is a complete sentence, so we extract it. \\
Answer: The proposed method improved performance at the 5\% significance level for the LSTM- and BERT-based models, and at the 10\% significance level for the conventional hybrid model. \\
 \\
\# Target Sentence \\
Sentence: \{claim\} \\
 \\
Analyse the sentence and extract the embedded sentence being reported. If the sentence is not a reporting clause, or if the reported content is not a complete sentence, return the original sentence. Give your reasoning first, then provide the extracted sentence. Use the format below. \\
Reasoning: <your reasoning> \\
Answer: <extracted sentence> \\
\end{prompt}

\begin{prompt}[title={\containshedging\hfill{}(\S\ref{subsec:flags})}]
\# Task \\
Determine whether the target sentence includes explicit hedging. \\
 \\
\# Instructions \\
* A sentence includes explicit hedging if it contains words or phrases that directly express uncertainty, possibility, approximation, limited generality, or reduced commitment, e.g. "may", "might", "tend to", "likely", "necessarily", "probably", "could", "would", "should", "can". \\
* Do not infer hedging unless it is explicitly signalled by the wording. \\
 \\
\# Target sentence \\
Sentence: \{claim\} \\
 \\
Does the target sentence include explicit hedging? State the hedging expression in the sentence first and then answer with Yes or No. The output should be in the format below. \\
Hedging expression: <exact word(s) or phrase(s), or "none"> \\
Answer: Yes/No \\
\end{prompt}

\begin{prompt}[title={\containsvague\hfill{}(\S\ref{subsec:flags})}]
\# Task \\
Determine whether the target sentence includes an under-specified adjective or adverb. \\
 \\
\# Instructions \\
* If there is an adjective or adverb in the sentence, consider whether it is a valid question to ask "how <adjective/adverb>?" If so, then the adjective/adverb is under-specified, answer Yes. \\
* If there is no adjective or adverb in the sentence, then answer No. \\
* When there are both specified and under-specified adjectives/adverbs in the sentence, count it as under-specified. \\
* Example of valid question: "how significant?", "how good?", "how efficient?" \\
* Example of non-valid question: "how true?", "how better?", "how best?", "how middle?", "how average?" \\
 \\
* More examples: \\
Sentence: "Furthermore, we observe significant discrepancies when applying the models to different domains." \\
Question: "how significant?" \\
Valid: Yes \\
Answer: Yes \\
 \\
Sentence: "The same is true here." \\
Question: "how true?" \\
Valid: No \\
Answer: No \\
 \\
Sentence: "In order to guarantee the annotation consistency, people turn to predefined annotation guidelines according to specific tasks or applications." \\
Question: "how predefined?", "how specific?" \\
Valid: No, Yes \\
Answer: Yes \\
 \\
Sentence: "The boxplots confirm the strength of CRF over the other models." \\
Question: None \\
Valid: NA \\
Answer: No \\
 \\
\# Target sentence \\
Sentence: \{claim\} \\
 \\
Does the target sentence include an under-specified adjective or adverb? State the question you would ask, determine whether it is valid, and then answer with Yes or No. The output should be in the format below. \\
Question: <question(s) you would ask, or "None"> \\
Valid: <Yes/No/NA for each question> \\
Answer: Yes/No \\
\end{prompt}

\endgroup

\label{app:prompt_templates}

\end{document}